\documentclass[conference]{IEEEtran}

\RequirePackage[l2tabu, orthodox]{nag}

\usepackage{cite}
\usepackage{amsmath,amssymb,amsfonts}
\usepackage{algorithmic}
\usepackage{threeparttable}
\usepackage[pdftex]{graphicx}
\graphicspath{{./figures/}{./resources/}}
\DeclareGraphicsExtensions{.pdf,.png,.jpg,.jpeg}

\usepackage{tabularx}
\usepackage{textcomp}
\usepackage{xcolor}
\usepackage[normalem]{ulem}
\usepackage{multicol}
\usepackage{hyperref}
\usepackage{multirow}
\usepackage{graphicx}
\usepackage{subcaption}
\usepackage{tikz}
\usetikzlibrary{decorations.pathreplacing}
\usepackage{array}
\usepackage{booktabs}
\usepackage{caption}
\usepackage{soul}
\usepackage{longtable}

\usepackage{booktabs, makecell}

\usepackage{xspace}

\begin{document}

\title{Exploring Kolmogorov-Arnold Networks for Interpretable Time Series Classification}

\author{\IEEEauthorblockN{%
    Irina Barašin, Blaž Bertalanič, Mihael Mohorčič and Carolina Fortuna
}\IEEEauthorblockA{%
    Department of Communication Systems, Jo\v{z}ef Stefan Institute\\ Jamova ulica 39, 1000 Ljubljana, Slovenia\\
}
irina.barasin@gmail.com, \{blaz.bertalanic, miha.mohorcic, carolina.fortuna\}@ijs.si%
\\
}

\maketitle
\thispagestyle{plain}
\pagestyle{plain}

\begin{abstract}

Time series classification is a relevant step supporting decision-making processes in various domains, and deep neural models have shown promising performance in this respect.
Despite significant advancements in deep learning, the theoretical understanding of how and why complex architectures function remains limited, prompting the need for more interpretable models. Recently, the Kolmogorov-Arnold Networks (KANs) have been proposed as a more interpretable alternative to deep learning. While KAN-related research is significantly rising,  to date, the study of KAN architectures for time series classification has been limited. 
In this paper, we aim to conduct a comprehensive and robust exploration of the KAN architecture for time series classification  utilising 117 datasets from UCR benchmark archive, from multiple different domains.
 More specifically, we investigate a) the transferability of reference architectures designed for regression to classification tasks, b) identifying the hyperparameter and implementation configurations for an architecture that best generalizes across 117 datasets, c) the associated complexity trade-offs and d) evaluate KANs interpretability. Our results demonstrate that (1) the Efficient KAN outperforms MLPs in both performance and training times, showcasing its suitability for classification tasks. (2) Efficient KAN exhibits greater stability than the original KAN across grid sizes, depths, and layer configurations, especially when lower learning rates are employed. (3) KAN achieves competitive accuracy compared to state-of-the-art models such as HIVE-COTE2 and InceptionTime, while maintaining smaller architectures and faster training times, highlighting its favorable balance of performance and transparency. (4) The interpretability of the KAN model, as confirmed by SHAP analysis, reinforces its capacity for transparent decision-making.
\end{abstract}

\section{Introduction}

Time series classification is a relevant step supporting decision-making processes in various domains. For instance, in healthcare time series classification enables the diagnosis and monitoring of conditions by analyzing patterns in physiological signals such as electrocardiograms or brain activity \cite{wang2022systematic}. In finance, it enhances risk management and fraud detection by classifying trading behaviors and transaction patterns \cite{devaki2014credit}. By classifying activities based on sensor data in human activity recognition, it supports personalized recommendations and safety monitoring \cite{yang2015deep}. In remote sensing, it contributes to environmental monitoring and land use classification by categorizing satellite or aerial imagery data over time \cite{gomez2016optical}. In these fields, accurate time series classification plays a central role in advancing predictive and analytic capabilities that drive critical decisions.

Unlike time series forecasting, which aims to predict future data points, time series classification aims to provide accurate interpretation and categorization of temporal data sequences by assigning a categorical label or a class label to each sequence \cite{bagnall2024hands}. Following a  well-accepted survey \cite{ismail2019deep} that revealed the lack of investigation of deep learning techniques for time series classification, significant progress has been achieved in related research with hundreds of new works, some of them surpassing the non-deep state of the art \cite{middlehurst2024bake}. According to the most recent and extensive benchmarking work on more than 100 datasets from the UCR archive, the most extensive benchmark for time series classification, the state of the art F1 score of 0.886 is achieved by Hive-Cote2 \cite{bagnall2015time} \cite{middlehurst2021hive}, a large ensemble with considerable training time. The non-deep and non-hybrid classification methods that are more explainable at both feature engineering and decision making steps achieve F1 of up to 0.869 while the deep architectures reach around 0.88 F1 score. Foundation models for time series, based on transformers, are yet to be studied on cross-domain benchmarks such as UCR~\cite{liang2024foundation}.

Despite the disruptive advancements introduced by breakthroughs in machine vision and natural language processing over the last decades, the theoretical understanding of why and how complex deep architectures function has lagged behind~\cite{bachmann2024scaling}. This has prompted researchers to step back and reconsider fundamental and mathematically simpler architectures, such as MLPs~\cite{teney2024neural}. Furthermore, the influence of the academic scientific community in this new era of AI is decreasing~\cite{Ahmed2023}, while non-AI scientific communities are concerned with the interpretability of deep models~\cite{bachmann2024scaling}.
Recently, the Kolmogorov-Arnold Networks (KANs)~\cite{liu2024kan} have been proposed as addressing the limitations of traditional neural networks. KANs have   demonstrated to be interpretable, even enabling symbolic regression, and have comparable performance with MLPs on small scale and science tasks with shallower architectures. Additional advantages have been explored for data fitting and solving partial differential equations \cite{liu2024kan2}.

The main critique of the original KAN work is  concerned with the fairness of the comparison, triggering substantial research into better understanding of their overall performance beyond the initial small scale scientific tasks. Subsequently it has been shown that their performance advantage in non-scientific ML tasks such as vision, natural language and audio processing does not hold \cite{yu2024kan}. Additional work reveals their reduced effectiveness on functions with noise \cite{shen2024reduced} while other works investigate the suitability of replacing the spline functions with wavelets \cite{bozorgasl2024wav}. The performance of KANs vs MLPs on graph learning tasks is compared in \cite{bresson2024kagnns}. Their preliminary results reveal that while KANs are on-par with MLPs in classification tasks, they seem to have a clear advantage in the graph regression tasks.

While KAN-related research is significantly rising, with new scientific works published almost weekly, the study of KAN architectures for time series has been limited to date. Very early forecasting studies encompass Temporal KANs (TKANs)~\cite{genet2024tkan}, Temporal Kolmogorov Arnold Transformer \cite{genet2024temporal} and mixture-of-experts \cite{han2024kan4tsf} for various domains from traffic and weather to satellite traffic  \cite{vaca2024kolmogorov} and integrating convolutional layers with KANs to improve time-series forecasting~\cite{livieris2024c}.
Furthermore, with respect to time series classification, studies on KANs robustness are emerging  \cite{dong2024kolmogorov}.

In this paper, we aim to conduct a comprehensive and robust exploration of the KAN architecture for time series classification on the UCR benchmark. More specifically, we look at a) how the existing architectures for forecasting \cite{vaca2024kolmogorov} transfer to classification, b) the hyperparameter and implementation influence on the classification performance in view of finding the one that performs best on the selected benchmark, c) the complexity trade-offs, and d) interpretability advantages. The contributions of this paper are as follows.
\begin{itemize}
    \item A study on the suitability of KAN architectures for classification tasks on the UCR benchmark consisting of 117 datasets. The study first investigates the feasibility of transferring existing architectures designed for regression followed by finding the most suitable  architecture for classification.
    \item A hyperparameter impact analysis of two KAN implementations, analyzing how variations in grid size, network depth, and node configurations impact classification performance. The analysis leads to finding the best performing KAN configuration that best generalizes across the 117 datasets.
    \item The performance and computational complexity comparison of the original KAN, the Efficient KAN implementation, and Multi-Layer Perceptrons (MLPs) on time series classification tasks revealing Efficient KAN's superior stability across grid sizes, depths, and layer configurations.
    \item We confirm KAN’s interpretability by diving deeper into the learnt feature importance and activations functions vs SHAP importance.
\end{itemize}

The paper is structured as follows. We discuss related work in Section \ref{sec:related}. Section \ref{sec:statement} outlines the problem statement, followed by Section \ref{sec:methodology} detailing  methodological aspects. A comprehensive analysis of the results is presented in Section \ref{sec:results}. Lastly, the paper concludes with Section \ref{sec:conclusion}.

\section{Related work}
\label{sec:related}
Kolmogorov-Arnold Networks (KANs) emerged as an innovative alternative to traditional Multi-Layer Perceptrons (MLPs), inspired by the Kolmogorov-Arnold representation theorem. KANs were shown to outperform MLPs in both accuracy and interpretability on small-scale scientific tasks. Their design, which models complex functions and patterns with fewer parameters, demonstrated potential to aid in mathematical and physical discoveries \cite{liu2024kan}. Building on these findings, subsequent research \cite{liu2024kan2} further developed KANs to bridge the gap between artificial intelligence and scientific research. By allowing KANs to identify relevant features, reveal modular structures, and discover symbolic formulas, KAN 2.0 \cite{liu2024kan2} introduced a bidirectional approach that not only incorporates scientific knowledge into KANs but also enables KANs to extract interpretable scientific insights from data. However, KANs have been shown to be highly sensitive to noise, prompting the introduction of oversampling and denoising techniques, such as kernel filtering with diffusion maps, to mitigate noise effects\cite{shen2024reduced}. To improve the handling of noisy data, architectural extensions have additionally explored the replacement of splines with wavelets \cite{bozorgasl2024wav}, enhancing KAN’s robustness and adaptability in broader applications.

KANs and MLPs were also investigated beyond scientific datasets in various domains in a controlled study with consistent parameters and FLOPs \cite{yu2024kan}. While MLPs outperformed KANs in most areas, KANs retained a distinct advantage in symbolic formula representation due to B-spline activation functions. Replacing MLPs' activations with B-splines improved their performance, suggesting KAN-inspired enhancements for MLPs. However, KANs faced memory stability issues in continual learning, requiring specialized tuning or hybrid models. In graph learning tasks \cite{bresson2024kagnns}, experiments on node classification, graph classification, and graph regression datasets indicated that KANs are on par with MLPs in classification but exhibit an advantage in graph regression tasks.

In a real-world satellite traffic forecasting task, KANs achieved comparable or superior accuracy to MLPs while using fewer parameters, showcasing their potential in predictive analytics \cite{vaca2024kolmogorov}. This application framed forecasting as a supervised learning problem with specific input-output mappings across time steps, using a GEO satellite traffic dataset. The success of KANs in this setting prompted specialized variants, such as Temporal KAN (T-KAN) and Multivariate Temporal KAN (MT-KAN) \cite{xu2024kolmogorov}. T-KAN targets univariate time series, capturing nonlinear relationships with symbolic regression, while MT-KAN models dependencies between multiple variables for improved accuracy in multivariate settings. Further advancements, such as Temporal Kolmogorov-Arnold Networks (TKANs), incorporated LSTM-inspired memory layers, excelling in tasks such as cryptocurrency trading volume forecasting. Temporal Kolmogorov-Arnold Transformer (TKAT) added self-attention mechanisms, outperforming conventional transformers in interpretability and precision in multivariate time series forcasting \cite{genet2024tkan, genet2024temporal}. Signature-Weighted KANs (SigKAN) extended these innovations by integrating path signatures, making them robust for market volume prediction \cite{inzirillo2024sigkan}. %
The Reversible Mixture of KAN Experts (RMoK) model, another KAN-based approach, introduced a mixture-of-experts structure to assign variables to KAN experts, achieving strong performance in time series forecasting tasks by leveraging temporal feature weights to explain data periodicity \cite{han2024kan4tsf}.

Although T‑KAN, MT‑KAN, TKAN, TKAT and SigKAN extend the Kolmogorov–Arnold mapping, they integrate forecasting‑specific modules (e.g. temporal kernels, attention masks) that diverge from the original design. As a result, they are unfit for classification without stripping or redesigning these components and retraining under a classification objective.

To enhance time series classification robustness, hybrid models that combine KAN and MLP architectures were further explored \cite{dong2024kolmogorov}. The study employed Efficient KAN as the primary implementation of KAN rather than the original implementation. Two hybrid configurations, KAN MLP (KAN with an MLP as the final layer) and MLP KAN (MLP with a KAN as the final layer), were tested across the UCR datasets, and both hybrids achieved performance comparable to traditional KAN and MLP models. Notably, MLP KAN demonstrated increased resilience against adversarial attacks. This robustness is attributed to the lower Lipschitz constant of KAN layers, suggesting that combining KAN with MLP structures can strengthen adversarial resistance in neural networks, thus opening possibilities for more secure and reliable models. Rather than focusing on robustness and hybrid architecture, this study aims at understanding in depth the performance and interpretability of KANs for time series classification.

\section{Problem Statement}
\label{sec:statement}
In this paper we analyze the potential of KAN for solving univariate time-series classification problems. A univariate time series \( X \) consists of a sequence of observations $[x_{1}, x_{2}, \ldots, x_{T}]$, where each $x_t, t \in \{1, 2, ..., T\}$ represents a value observed at time \( t \) representing an ordered set of real values \cite{ismail2019deep}. 

\begin{figure}[h]
    \centering
    \includegraphics[width=\linewidth]{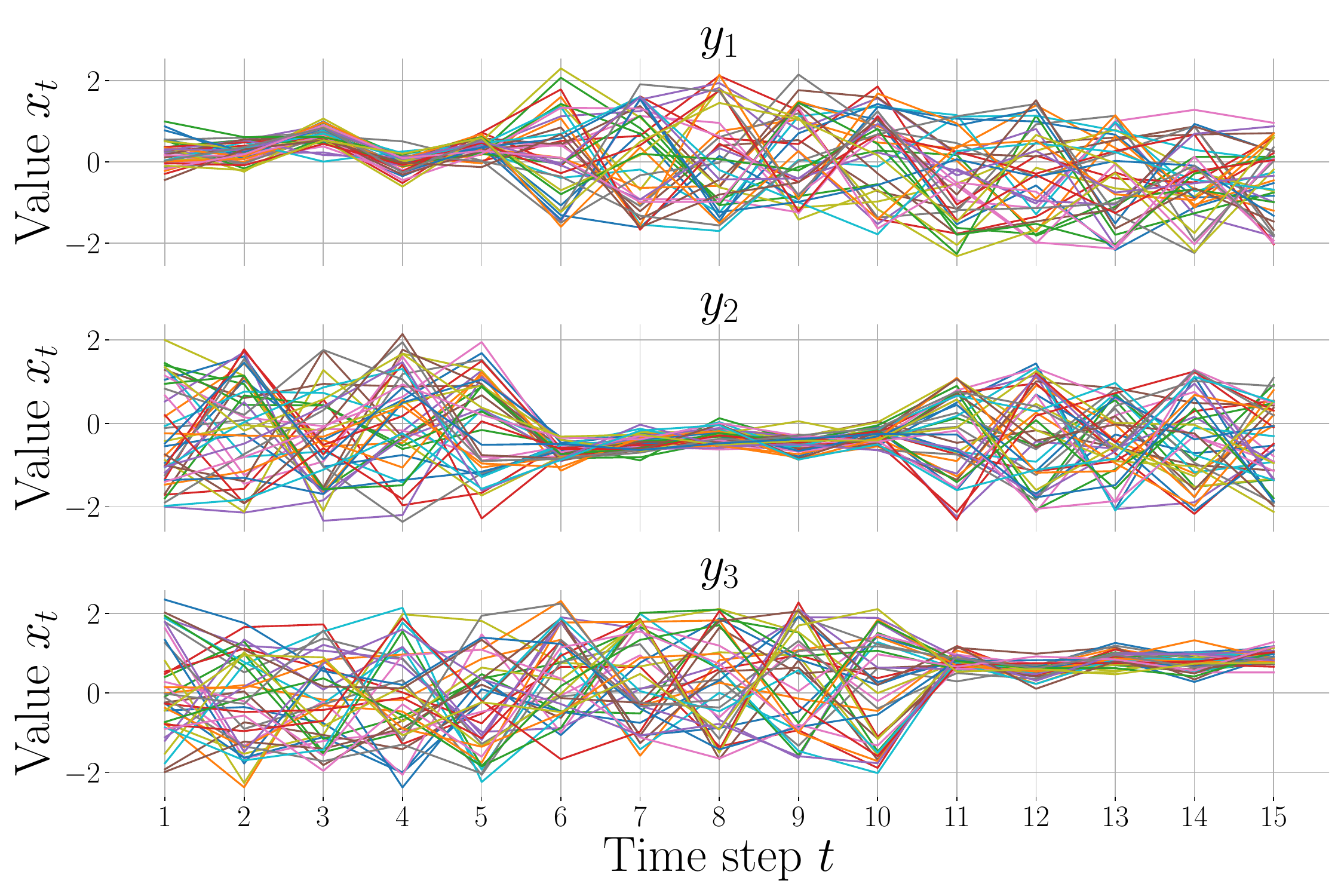}
    \caption{Example of time series data plotted by class, with time steps on the x-axis and observed values $x_t$ on the y-axis. Figure depicts SmoothSubspace, one of the 117 datasets from the UCR repository.}
    \label{fig:ts_classes}
\end{figure}

A class label \( Y \) is a categorical variable associated with a time series \( X \), indicating the class that the time series belongs to. The number of distinct classes is denoted by the cardinality $C$ of  \( Y \), meaning \( Y \) can take a value from the set $\{y_1, y_2, \ldots, y_C\}$. When $C=2$, this corresponds to a binary classification problem, while $C>2$ represents a multiclass classification problem. As shown in Fig. \ref{fig:ts_classes}, where \( C = 3 \), each plot corresponds to a different class, illustrating the distinct patterns for \( y_1 \), \( y_2 \), and \( y_3 \).

A dataset $D = \{(X_1, Y_1), (X_2, Y_2), \ldots, (X_N, Y_N)\}$ consists of pairs $(X_i, Y_i) $, where $X_i, i \in \{1, 2, 3..., N\}$ represents a univariate time series, and $Y_i$ is the corresponding class label.

The task of time series classification involves training a classifier on a dataset \(D\) to map each input \(X_i\) to a probability distribution across the possible class labels \cite{ismail2019deep}.

\begin{equation}
f(D) \approx Y 
\end{equation} 

Specifically, instead of assigning a single class label to each time series, the classifier outputs a probability distribution across all possible classes. That is, for each time series \( X_i \), the classifier predicts a vector of probabilities \( P(Y_i = y_c \mid X_i) \) for each class \( c \in \{1, 2, \dots, C\} \).

The classification function $f$ can be realized through a multitude of exiting approaches \cite{middlehurst2024bake}, however we investigate the recently proposed Kolmogorov Arnold Network (KAN) \cite{liu2024kan} promising increased interpretability and potential computational benefits compared to standard Multilayer Perceptron (MLP).

\subsection{KAN for time series classification}
\label{sec:prob_stat_kan}

\begin{figure}[!b]
    \centering
    \includegraphics[width=\linewidth]{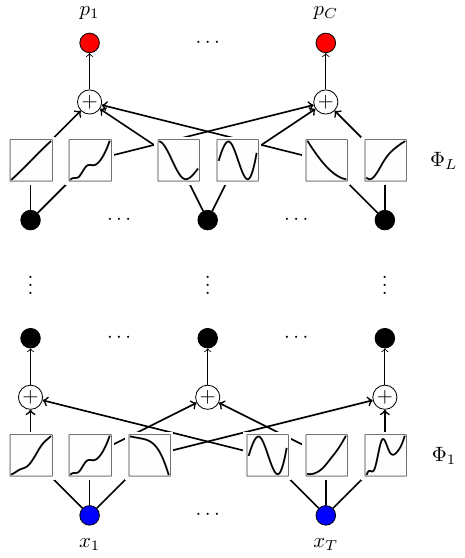}

 \caption{KAN architecture.}
    \label{fig:kan_arch}
\end{figure}

The architecture of a KAN is depicted in Fig. \ref{fig:kan_arch}. It can be seen from the figure that, for an input instance \( X_i \), the network outputs \( Y_i \) label assignment probability vector. \( Y \) is learnt from \( D \) as a composition of $L$ appropriate univariate functions $\Phi$, as depicted in Fig. \ref{fig:kan_arch} and formalized in Eq.~\ref{eq:eq_phi_composition1}, where $L$ stands for the number of layers in the network. 

Each layer of a KAN is represented by a matrix where each entry is an activation function.
If there is a layer with $d_{in}$ nodes and its neighboring layer with $d_{out}$ nodes, the layer can be represented as a $d_{in} \times d_{out}$ matrix of activation functions: %

The structure of a KAN can be represented as $[n_1,…,n_{L+1}]$, where $L$ signifies the total number of layers in the KAN. A deeper KAN can be thus formulated through the composition of $L$ layers as:
\begin{equation}
    Y = \text{KAN}(X) = (\Phi_L \circ \Phi_{L-1} \circ \dots \circ \Phi_1)X.
\label{eq:eq_phi_composition1}
\end{equation}

\begin{equation}
\Phi = \{\phi_{q,p}\}, \; p = 1, 2, \dots, d_{in}, \; q = 1, 2, \dots, d_{out}.
\label{eq:eq_phi_composition2}
\end{equation}

 Unlike traditional MLPs, where activation functions are applied at the nodes themselves, KAN places them at the edges between the nodes. KAN employs the SiLU activation function in combination with B-splines to enhance its expressiveness, as per Eq.~\ref{activation_func}. This setup allows the edges to control the transformations between layers, while the nodes perform simple summation operations. 

\begin{equation}
\phi(x) = w_b silu(x) + w_s \, \text{spline}(x).
\label{activation_func}
\end{equation}

SiLU is defined as $\text{silu}(x) = x/(1+e^{-x})$. This activation function allows for smooth, non-linear transformations, which helps the network capture complex patterns in time series data more effectively.

B-splines are smooth polynomial functions of degree (order) $k$, which approximate data using control points. The order $k$ controls the smoothness of the spline. Commonly, $k=3$ is used for cubic splines. Each spline operates over a defined grid $G$, which divides the input space into smaller intervals. The grid is determined by a set of grid points that define the segments of the spline. Increasing the number of grid points increases spline resolution, enabling more fine grained approximation of the underlying univariate function. 

A B-spline of order $k$ requires $G+k$ basis functions to represent the spline over the grid. For each input (node in a layer), the evaluation of a B-spline of order $k$ thus involves computing $G+k-1$ basis functions and performing a weighted sum with the corresponding control points.
\begin{equation}
\text{spline}(x) = \sum_{i=0}^{G+k-1} c_i B_i(x)
\end{equation}

To better understand the performance of the evaluated models, it is crucial to explain their decision-making processes. Although black-box models achieve high accuracy, they often lack interpretability. KANs offer a shift in the interpretability paradigm by allowing visualization of the model structure along with its learned B-splines as shown in Fig.~\ref{fig:kan_arch}. Interpretability is further enhanced by model pruning that simplifies visual representation by removing less important connections within the model, which were determined through their significance scores. These scores identify critical connections, which highlight the most important features for the decision of the model. This approach provides insights into how KANs process data and make predictions by focusing on the most influential connections within the structure of the model.

\subsubsection{KAN learnable parameters}
The number of learnable parameters in a KAN layer is determined by its architecture, which includes contributions from the B-spline control points, shortcut path weights, B-spline weights, and bias terms. As discussed in \cite{yu2024kan}, the number of learnable parameters for each KAN layer is:
\begin{equation}
\text{Parameters} = (d_{in} \cdot d_{out}) \cdot (G + k + 3) + d_{out} ,
\end{equation}

\noindent where \(d_{in}\) and \(d_{out}\) represent the input and output dimensions of the layer. This is consistent with findings in \cite{liu2024kan}, which also emphasize the role of learnable control points and weights associated with the spline functions.

\subsubsection{KAN FLOPs computation}
\label{sec:kan_flops}
The number of FLOPs in a Kolmogorov–Arnold Network (KAN) is highly dependent on the specific implementation and the way the network is compiled on various processor architectures. Different hardware platforms optimize these computations differently, leading to variations in practical FLOPs performance. However, we adopt the theoretical FLOPs computation model as outlined in \cite{yu2024kan} for consistency and comparison purposes.

The total FLOPs for a KAN layer come from three parts: the B-spline transformation, the shortcut path, and the merging of the two branches. Using the De Boor-Cox iterative formulation for B-splines, the FLOPs for one KAN layer in the original KAN implementation is given by Eq.~\ref{eq:kan_flops} \cite{yu2024kan}:
\begin{equation}
\begin{split}
\text{FLOPs} = &\ \text{FLOPs of non-linear function} \cdot d_{\text{in}} \\
               &+ (d_{\text{in}} \cdot d_{\text{out}}) \cdot \left[ 9 \cdot k \cdot (G + 1.5 \cdot k) \right. \\
               &\left. + 2 \cdot G - 2.5 \cdot k + 3 \right].
\end{split}
\label{eq:kan_flops}
\end{equation}

FLOPs of one forward pass $FP$ through a network with uniform hidden layers of size $d_{in} \cdot d_{out} = M \cdot M$ are calculated as in Eq.~\ref{eq:kan_fp_flops}\cite{yu2024kan}:
\begin{equation}
\begin{split}
N_{FP}= &\ \text{FLOPs of non-linear function} \cdot T \\
                  &+ T \cdot M \cdot \big[ 9 \cdot k \cdot (G + 1.5 \cdot k) \\
                  &\quad\ + 2 \cdot G - 2.5 \cdot k + 3 \big] \\
                  &+ (L - 2) \cdot \big( \text{FLOPs of non-linear function} \cdot M \\
                  &\quad\ + M^2 \cdot \big[ 9 \cdot k \cdot (G + 1.5 \cdot k) \\
                  &\quad\quad\ + 2 \cdot G - 2.5 \cdot k + 3 \big] \big) \\
                  &+ \text{FLOPs of non-linear function} \cdot M \\
                  &+ M \cdot C \cdot \big[ 9 \cdot k \cdot (G + 1.5 \cdot k) \\
                  &\quad\ + 2 \cdot G - 2.5 \cdot k + 3 \big].
\end{split}
\label{eq:kan_fp_flops}
\end{equation}

\subsection{ Comparison of KAN to MLP for time series classification}
A Multilayer Perceptron (MLP) is  a very well known type of feed-forward neural network widely used for time series classification. %

 KANs and MLPs share several architectural principles but differ significantly in how they implement non-linearity and function approximation. Both are fully connected neural networks in which each layer's nodes are densely connected to the next. The input layer in each architecture corresponds to the length of the input time series, and the output layer produces a probability distribution over class labels, allowing for multiclass classification.

However, their treatment of activation differs notably. In MLPs, activation functions are applied at the neurons (i.e. nodes). In KANs, activation functions are applied on the edges between nodes.
Regarding activation functions, the two architectures also use different approaches. MLPs commonly use standard functions such as ReLU, defined as $\text{ReLU}(z) = \max(0, z)$. KANs employ the SiLU (Sigmoid Linear Unit) activation in combination with B-splines.

\subsubsection{MLP learnable parameters}
The number of learnable parameters in an MLP is determined by the connections between neurons across layers. For a fully connected layer with $d_{in}$ input neurons and $d_{out}$ output neurons, the number of learnable parameters \cite{yu2024kan} is:
\begin{equation}
Parameters = (d_{in} × d_{out})+ d_{out}.
\end{equation}

\subsubsection{MLP FLOPs computation}
In a fully connected MLP, each connection between two neurons $wx+b$ performs a weighted sum $w \cdot x$ and adds bias $b$, resulting in 1 multiplication and 1 addition. Thus, each connection requires 2 FLOPs. 

For a neuron with $d_{\text{in}}$ inputs, calculating the output requires $2 \times d_{\text{in}}$ FLOPs (for weights) and 1 FLOP for the bias, resulting in a total of $2 \times d_{\text{in}} + 1$ FLOPs. If there are $d_{\text{out}}$ output neurons, the total FLOPs for the layer is $d_{\text{out}} \times (2 \times d_{\text{in}} + 1)$.

We consider fully connected MLP which has input layer with $T$ neurons, $K$ hidden layers, with $M$ neurons in each, and $C$ neurons in output layer. FLOPs for forward propagation are therefore calculated as: 
\begin{equation}
N_{FP} = (M + 2 M T) + (K-1)(M + 2 M^2) + (M + 2 M C).
\label{eq:mlp_flops}
\end{equation}

\section{Methodology}
\label{sec:methodology}

In this section we elaborate on the methodology adopted for this study\footnote{https://github.com/irina-b1/KAN\_TS}. First we provide considerations on data and preprocessing, followed by reference model configurations, training process, hyperparamater, complexity and interpretability analysis.

\subsection{Data and preprocessing}
\label{sec:data_methodology}
In this study, the UCR (University of California, Riverside)~\cite{dau2019ucr} dataset archive time series classification benchmark is utilized , which contains a total of 128 univariate datasets from diverse domains, ranging from ECG signals and motion capture data to spectrographs and simulated control systems.

The datasets were selected based on their completeness, since some datasets include time series instances containing missing values. To ensure that an adequate amount of data remained for both the training and testing phases, these datasets were excluded, resulting in total of 117 datasets used.  For each dataset, the Appendix provides the number of training and test instances, the length of time series, and the number of classes.

 Table~\ref{tab:datasets_diversity}   further shows descriptive statistics for 117 datasets. Each row represents a statistical measure: minimum, maximum, mean, median and standard deviation. The columns correspond to the number of training samples, the number of test samples, the length of each time series and the number of classes across the 117 UCR datasets.

\begin{figure*}[h]
    \centering
    \includegraphics[width=\linewidth]{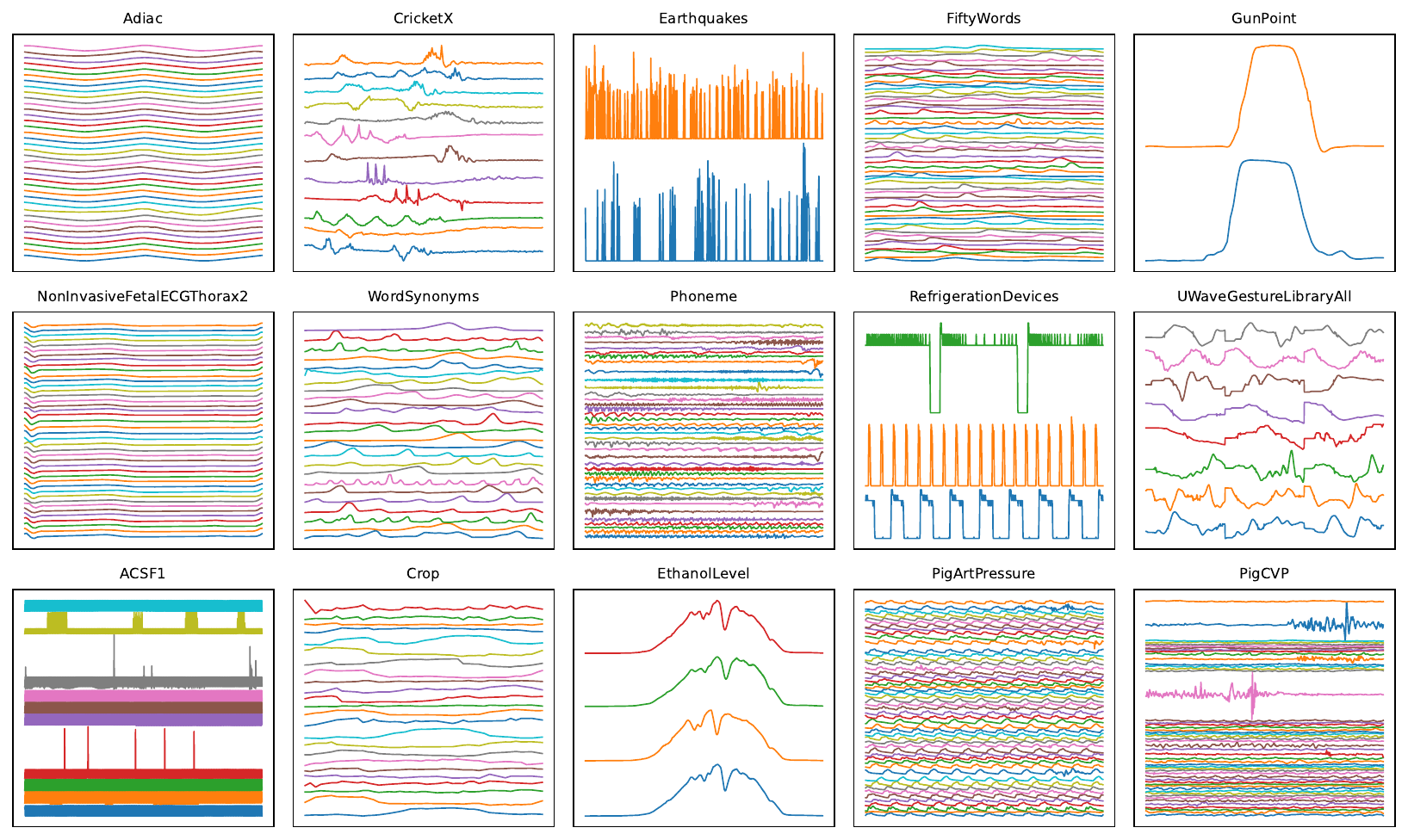}
    \caption{ 15 datasets showing one representative time series per class. Distinct colors indicate different classes}
    \label{fig:ucr_samples}
\end{figure*}

In terms of size, for training specifically, the datasets vary significantly, ranging from 16 up to 8,926 time series instances. The input lengths of these time series also showed considerable diversity, ranging from 15 to 2,844 time steps per instance in different datasets. Furthermore, the datasets included varying numbers of classes for classification tasks, with scenarios ranging from binary classifications with 2 classes to multiclass problems with up to 60 distinct classes.  To illustrate the diversity in shape and class structure, we visualize 15 randomly selected datasets in Fig.~\ref{fig:ucr_samples} , where each subplot displays one representative instance per class, with distinct colors denoting different classes.

The first step in data preprocessing was normalization, using `StandardScaler` from the scikit-learn library. Each time series \( X \) in the dataset \( D \) was standardized by applying the transformation \( x_{t,\text{scaled}} = \frac{x_t - \mu_t}{\sigma_t} \), where \( \mu_t \) and \( \sigma_t \) are the mean and standard deviation for the value at time \( t \) across all time series in \( D \). This ensures that each time point \( t \) has a mean of 0 and a standard deviation of 1, allowing for consistent scaling across the dataset. This normalization is crucial for ensuring that the input features are on a comparable scale, which helps in efficient training.

\textbf{Data Availability Statement.}  The data that support the findings of this study are available in the UCR Archive at https://doi.org/10.1109/JAS.2019.1911747, and are also available in the public domain at https://www.timeseriesclassification.com/.

\begin{table}[t!]
\centering
\begin{tabular}{|l|r|r|r|r|}
\hline
\textbf{Statistic} & \textbf{Train} & \textbf{Test} & \textbf{Length} & \textbf{Class} \\
\hline
Min     & 16     & 20      & 15    & 2  \\
Max     & 8 926  & 16 800  & 2 844 & 60 \\
Mean    & 496.72 & 1 086.72 & 537.10 & 8.26 \\
Median & 181 & 343 & 301 & 3 \\
StdDev  & 1 155.16 & 2 080.59 & 583.14 & 12.26 \\
\hline
\end{tabular}
\caption{Summary statistics of the 117 UCR archive time series datasets}
\label{tab:datasets_diversity}
\end{table}

\subsection{Training setup}
For training, each dataset was partitioned into a training
dataset and a validation dataset in an 80:20 ratio. The test datasets are provided separately by default.  To ensure robustness and mitigate the effects of random initialization, each model was trained on 5 different random seeds. The presented results are averages of these 5 runs across all 117 datasets. The models were trained for 500 epochs using batch size 16.

The training process across all models used the Adam optimizer.  The baseline architectures from \cite{vaca2024kolmogorov}  were trained with a learning rate of $0.001$. However, instead of the mean absolute error (MAE) loss function used for regression tasks in the original work, we employed the cross-entropy loss function to address the classification nature of our problem. Additionally, we introduced $L1$ regularization with a weight factor of $0.1$.

Similarly, in the baseline MLP models, we used the same Adam optimizer, learning rate ($0.001$), and cross-entropy loss function as in the KAN models, for the same reasons. However, instead of L1 we applied L2 regularization with a weight factor of 1, selected through empirical tuning.

The original KAN library \cite{liu2024kan} was modified to enable GPU support, allowing the experiments to leverage the computational power of GPUs rather than relying solely on CPU execution. Thus, all evaluations of each model were conducted on an NVIDIA A100 80GB GPU.

For performance evaluation, we use precision, recall and F1 score. The precision measures how many instances predicted as a certain class actually belong to that class, expressed as: $\text{Precision} = \frac{\text{TP}}{\text{TP} + \text{FP}}$, where TP (true positives) represents the number of correctly identified positive instances, and FP (false positives) refers to the number of instances incorrectly identified as positive. Recall measures how many instances of some class were correctly detected. It is given by: $\text{Recall} = \frac{\text{TP}}{\text{TP} + \text{FN}}$, where FN (false negatives) represents the number of actual positive instances that the model failed to identify as positive. The F1 score is expressed as: $F1 = 2 \times \frac{\text{Precision} \times \text{Recall}}{\text{Precision} + \text{Recall}}$, where higher values indicate better performance in balancing precision and recall (better classification).

\subsection{Reference architecture configurations}
\label{sec:model_config}

 Initially, we analyze the performance of KAN and MLP by adopting the regression designs from the extensive empirical study in \cite{vaca2024kolmogorov}. The goal is to gain insights into the  generalization capabilities and transfer potential of these architectures from regression to classification problems.

As discussed in Section \ref{sec:statement}, we employ two different models: KAN and MLP. Alongside the original implementation of KAN\footnote{github.com/KindXiaoming/pykan}, we also include its variant, Efficient KAN\footnote{github.com/Blealtan/efficient-kan}, which optimizes memory by reformulating the activation process, applying B-splines directly to inputs and combining them linearly, thus avoiding the need for large tensor expansions. It replaces input-based L1 regularization with weight-based L1, improving efficiency. The L1 regularization is now computed as mean absolute value of the spline weights \cite{efficient-kan}.

KAN and Efficient KAN are configured as [$T$, 40, 40, $C$], representing a dapth 3 network, where $T$ represents the length of the input time series, while $C$ denotes the number of classes for classification, as per Section \ref{sec:statement}.  Although this architecture appears to involve two layers between input and output, in KANs depth refers to the number of trainable connections between layers. Thus, although there are two hidden layers, the model has three learnable transformations. A depth 4 variant includes an additional hidden layer with 40 nodes. We retained the spline order $k = 3$ and a fixed grid size $G = 5$.

In contrast, the MLP architecture is configured as [$T$, 300, 300, 300, $C$], representing a depth 3 network. The depth 4 variant adds a hidden layer of $300$ nodes. 

\subsection{Hyperparameter impact analysis}
\label{sec:hyper_impact_methodology}
To provide deeper insights into the KAN architecture's performance and provide design guidelines, we conduct an analysis of hyperparameters and configuration parameters with the goal of finding the best model for time series classification on the UCR behcmark , using a curated subset of 117 UCR datasets. To assess hyperparameter effects, we systematically varied grid size ($G$), network depth ($L$) and layer width ($M$). Table~\ref{tab:hyperparameters_table}  summarizes these parameters, where the first column lists each hyperparameter and the second indicates the range of values evaluated. For each of these evaluations, the models were trained using different learning rates \(lr \in \{0.0001, 0.001, 0.01, 0.1, 1\}\). 

The first evaluation examined the effect of grid size, where \( G \in \{3, 5, 10, 15, 20\} \) while keeping depth and layer size parameters fixed. Specifically, \(L=3, M=40\). 

In the second evaluation, we varied the depth of the network by adjusting the number of layers \( L \) between 2 and 10, with constant grid \( G = 5 \) and layer size \( M = 40 \). 

In the final evaluation, the number of nodes per layer \( M \in \{5, 10, \dots 100\} \) was varied, while keeping the depth fixed at $L=3$ and grid size at $G=5$.

 In summary, to ensure robust and representative results, separate models were trained for each of the 117 UCR datasets, with input and output layer sizes matched to the the specific time series lengths and number of classes, respectively. Multiple hyperparameter settings were tested for each one, including different learning rates, depths, hidden layer sizes, and spline grid resolutions. Taken together, this resulted in the development and evaluation of several tens of thousands of individual models across the full range of datasets and configurations within the UCR archive.

\begin{table}[ht]
\centering

\label{tab:kan_hyperparams}
\begin{tabular}{ll}
\toprule
\textbf{Hyperparameters} & \textbf{Values} \\
\midrule
Grid size  & $G \in \{3, 5, 10, 15, 20\}$ \\
Network depth & $L \in \{2, 3, \ldots, 10\}$ \\
Hidden layer size & $M \in \{5, 10, \ldots, 100\}$ \\
Learning rate & \(lr \in \{0.0001, 0.001, 0.01, 0.1, 1\}\) \\
Random seed & $ \{0, 1, 2, 5, 42\}$ \\
Batch size & $16$ \\
Epochs & $500$ \\
\bottomrule
\end{tabular}
\caption{ Hyperparameters utilised for impact analysis for KAN and Efficient KAN models.}
\label{tab:hyperparameters_table}
\end{table}

\subsection{Complexity analysis}
\label{sec:complexity_analysis}
In this section we evaluate resource consumption by calculating the average FLOPs over the 117 UCR datasets for one prediction, and the theoretical energy consumption (TEC) per prediction. FLOPs were computed  theoretically using Eqs. \ref{eq:kan_fp_flops} and \ref{eq:mlp_flops} outlined in Section \ref{sec:statement}. As per Section~\ref{sec:data_methodology}, \( T \in \{16, \ldots, 8,926\} \), \( C \in \{2, \ldots, 60\} \). %

We estimated TEC as \( \text{TEC} = \frac{\text{FLOPs}}{\text{FLOPS/Watt}} \) \cite{subramaniam2013trends}, where FLOPS/Watt represents the number of floating-point operations executed per second per watt. As the experiments were conducted on an NVIDIA A100 80GB PCIe GPU, its theoretical power consumption is 65 GFLOPS/Watt, for float32 operations, which were used for calculation.

\subsection{Interpretability analysis}
\label{sec:interpret_methodology}

We demonstrate the interpretability of KANs on architecture of shape [15, 15, 15, 3], that is amenable to   visualization, with a grid size G=5, learning rate $lr=1$, and L1 regularization of 0.01. The architecture and hyperparameters were iteratively optimized to ensure the model's performance was competitive with MLPs and HiveCote 2.0, ensuring that the model achieved comparable results while being significantly less computationally complex. The interpretability results are showcased on the SmoothSubspace dataset from the UCR repository, which has a length of 15 and 3 output classes. However, the methods are equally effective across all datasets in the UCR repository.

 Furthermore, to validate the inherent interpretability of KAN and to extract any insight from the MLP, global SHapley Additive exPlanations (SHAP) values~\cite{lundberg2017unified}  were used. 
SHAP offers a model-agnostic framework for global interpretability by quantifying the average contribution of each feature to the model's predictions across the entire dataset. The contribution of a feature $x_i$ is represented by its Shapley value, $\phi_i$, which is computed based on cooperative game theory. For a model the prediction for an instance $X$  is expressed as:

\begin{equation}
    SHAP = \phi_0 + \sum_{i=1}^M \phi_i,
\end{equation}

\noindent where $\phi_0$ is the model’s baseline expected value across the dataset, $\phi_i$ is the Shapley value of feature  $x_i$, and M represents the total number of features. By aggregating these values, SHAP enables insights into the global decisions of the model. In addition, SHAP is also utilized to compare the feature importance of KANs to the MLP architecture.

\begin{table*}[h!]
\centering
\begin{tabular}{|l|l|c|c|c|c|c|c|c|}
\hline
\multirow{2}{*}{\textbf{Model}}  & \multirow{2}{*}{\textbf{Configuration}} & \multicolumn{2}{c|}{\textbf{Precision}} & \multicolumn{2}{c|}{\textbf{Recall}} & \multicolumn{2}{c|}{\textbf{F1 score}} & \multirow{2}{*}{\textbf{Training Time (s)}}\\
\cline{3-8}
& & \textbf{Mean} & \textbf{StdDev} & \textbf{Mean} & \textbf{StdDev} & \textbf{Mean} & \textbf{StdDev} & \\ \hline

MLP (3-depth)           & [300, 300, 300]           & \textbf{0.73} & 0.20 & 0.66           & 0.23 & 0.64 & 0.25 & 109.50\\ 
MLP (4-depth)           & [300, 300, 300, 300]      & 0.73          & 0.20 & 0.65           & 0.20 & 0.62 & 0.26 & 105.53\\ 
KAN (3-depth)           & [40, 40], \( G = 5 \)     & 0.38 & 0.17    & 0.33 & 0.21          & 0.30 & 0.21  & \textbf{44.99}\\ 
KAN (4-depth)           & [40, 40, 40], \( G = 5 \) & 0.37          & 0.11 & 0.32           & 0.20 & 0.29 & 0.20 & 56.88\\ 
Efficient KAN (3-depth) & [40, 40], \( G = 5 \)     & 0.72          & 0.20 & \textbf{0.70}  & 0.21 & \textbf{0.70} & 0.22 & 90.80\\ 
Efficient KAN (4-depth) & [40, 40, 40], \( G = 5 \) & 0.71           & 0.20 & 0.70          & 0.22 & 0.69 & 0.22  & 92.71\\  
\hline
\end{tabular}
\caption{Performance of reference regression architectures introduced in \cite{vaca2024kolmogorov}.}
\label{tab:model_performance}
\end{table*}

\section{Results}
\label{sec:results}
This section presents a comprehensive analysis of the results through classification performance of reference architectures, computational complexity comparison, hyperparameter impact analysis, and interpretability evaluation.

\subsection{Classification performance analysis of reference architectures}
\label{sec:class_performance}

Table \ref{tab:model_performance} presents the results of the classifiers with rows representing different models and their configurations. The first two rows provide the results for MLP models and the subsequent four rows for KAN and Efficient KAN models. The first column lists the type of model, while the second column details the specific configuration of each model, including the nodes per layer and grid size for KAN models. The subsequent columns provide the performance metrics for each model, split into three main categories: precision, recall, and F1 score. Each of these categories is further divided into two subcolumns that show the mean and standard deviation (StdDev) of the results across multiple (i.e. 5) runs. The last column presents average time, measured in seconds, for training one model. 

The MLP models demonstrate relatively high and consistent performance across the precision, recall, and F1 score metrics for both depths. For the 3-layer MLP, the mean precision is  0.73 with a standard deviation of 0.20, and the 4-layer model exhibits the same mean and standard deviation for precision. In terms of recall, the mean values are very similar, at 0.66 for the 3-layer model and 0.65 for the 4-layer model, with standard deviations of 0.23 and 0.20, respectively. This indicates that increasing the depth has little effect on the model’s ability to identify relevant instances. Additionally, both models maintain comparable F1 score means of 0.64 (3 layers) and 0.62 (4 layers), although the standard deviation increases slightly from 0.25 to 0.26, suggesting a minor decrease in consistency for the F1 score with the deeper model.

The original KAN implementations show a substantial drop in performance compared to the MLP models. The 3-depth KAN has a mean precision of 0.38, with a recall of 0.33 and an F1 score of 0.30. 
The 4-depth KAN model continues this trend with slightly lower performance, showing a mean precision of 0.37, recall of 0.32, and F1 score of 0.29. The further decline in performance suggests that adding depth does not improve generalization but may instead lead to overfitting, especially across multiple diverse datasets.
These results indicate that the original KAN struggles to match the classification performance of the MLP. While previous research demonstrated that KAN performed well for time series prediction and exhibited strong generalization abilities \cite{vaca2024kolmogorov}, here KAN's transferability to a different problem, i.e. classification, appears less successful. 

Lower performance results come with a significantly reduced training time of $\sim 45$ seconds for depth 3 KAN, due to the GPU acceleration, as described in Section \ref{sec:complexity_analysis}, suggesting a trade-off between performance and efficiency. However, the training time in depth 4 KAN increases to $\sim 57$ seconds, reflecting the impact of the added depth on computational cost.

\begin{table*}[t!]
\centering
\begin{tabular}{|l|l|c|c|c|c|}
\hline
\textbf{Model}  & \textbf{Configuration} & \textbf{Lernable Parameters} & \textbf{Theor. FLOPs}  & \textbf{TEC (Joules)} \\
\hline
MLP (3-depth)           & [300, 300, 300]           & 344 518&\textbf{688 420}& $\boldsymbol{1.008 \times 10^{-5}}$ \\ 
MLP (4-depth)           & [300, 300, 300, 300]      & 434 818 &868 720&  $1.336 \times 10^{-5}$ \\ 
KAN (3-depth)           & [40, 40], \( G = 5 \)     & \textbf{257 649}&6 146 993& $9.457 \times 10^{-5}$ \\ 
KAN (4-depth)           & [40, 40, 40], \( G = 5 \) & 275 289&6 566 993&  $10.103 \times 10^{-5}$ \\ 
Efficient KAN (3-depth) & [40, 40], \( G = 5 \)      &\textbf{257 649} & 6 146 993 & $9.457 \times 10^{-5}$ \\ 
Efficient KAN (4-depth) & [40, 40, 40], \( G = 5 \)  & 275 289 &6 566 993  & $10.103 \times 10^{-5}$ \\  
\hline
\end{tabular}
\caption{Comparison of performance and computational characteristics across the reference models}
\label{tab:resource_consumption}
\end{table*}

The models trained with the Efficient KAN implementation offer a notable improvement in performance compared to the original KAN implementation. The Efficient KAN with 3-depth demonstrates a significant boost in performance with a mean precision of 0.72, recall of 0.70, and F1 score of 0.70. 
Similarly, the 4-depth Efficient KAN, maintains high performance metrics with a mean precision of 0.71, recall of 0.70, and F1 score of 0.69. Improvements in comparison to KAN can be attributed to implementation changes, as discussed in Section \ref{sec:model_config}. 

The precision of Efficient KAN is slightly lower than that of the MLP models, which means that the model may produce more false positives. However, the Efficient KAN model compensates for this with higher recall, indicating that it is better at identifying true positives. As a result, the F1 score is also higher for Efficient KAN. This suggests that while Efficient KAN sacrifices a bit of precision, it achieves a better overall performance by improving recall and providing a more balanced model for classification.
Along with the increased performance, Efficient KAN maintains fast training times. The training time for the 3-depth model is approximately 91 seconds, with a slight increase to 93 seconds for the 4-depth model. This shows that while the original KAN implementation is less successful in transferring from time series prediction to classification tasks, the Efficient KAN not only achieves this transfer effectively but also generalizes better across different datasets, making it a more effective model overall.

\subsection{Complexity analysis for the reference models}
Table \ref{tab:resource_consumption} presents the computational complexity analysis for the MLP and KAN configurations. The first two columns, listing the model and configuration, are repeated from Table \ref{tab:model_performance}. The subsequent columns provide key complexity metrics: the third column displays the total number of lernable parameters in each model, while the fourth column reports the average theoretical FLOPs for one prediction across all used UCR datasets, based on model architectures discussed in Section \ref{sec:statement}. %
The last column presents the Theoretical Energy Consumption (TEC) in Joules, calculated based on the FLOPs and GPU efficiency, as per Section \ref{sec:complexity_analysis}. 

In the table we observe that MLPs, with approximately $344k$ and $434k$ learnable parameters for $3$-depth and $4$-depth, respectively, are the most computationally demanding models in terms of learnable parameters and training duration. Their theoretical FLOPs reach $688,420$ and $868,720$, while requiring $1.008 \times 10^{-5}$ and $1.336 \times 10^{-5}$ Joules of energy per single prediction, positioning them as the best-performing models in theoretical FLOPs and TEC (Total Energy Consumption).

Next, the table reveal that the original KAN implementation has noticeably fewer learnable parameters compared to the MLP models, consistent with the theoretical analysis in Section \ref{sec:prob_stat_kan}. Despite the reduction in parameters, they come with significantly higher theoretical FLOPs, with values of $6,146,993$ for depth 3 and $6,566,993$ for depth 4, due to B-spline computation, as per Eq. \ref{eq:kan_fp_flops}. The energy required for a single prediction also reflects this increase, amounting to $9.457 \times 10^{-5}$ Joules for depth 3 and $10.103 \times 10^{-5}$ Joules for depth 4. These results illustrate a trade-off, where KAN achieves a smaller parameter count but requires greater computational resources in terms of FLOPs and energy per prediction.

The last two rows for the Efficient KAN implementation show the same number of weights and theoretical FLOPs as the original KAN implementation, as the model architectures are identical. %
Despite higher energy consumption, Efficient KAN maintains fast training times. Along with the discussed increased performance, this balance of high performance and lower computational costs makes these models strong candidates for applications requiring both efficiency and effectiveness.

However, despite these improvements, MLP models remain far more energy-efficient, requiring 10 times less energy (TEC) per prediction than both KAN and Efficient KAN models.

\begin{figure}[h]
    \centering
    \includegraphics[width=\linewidth]{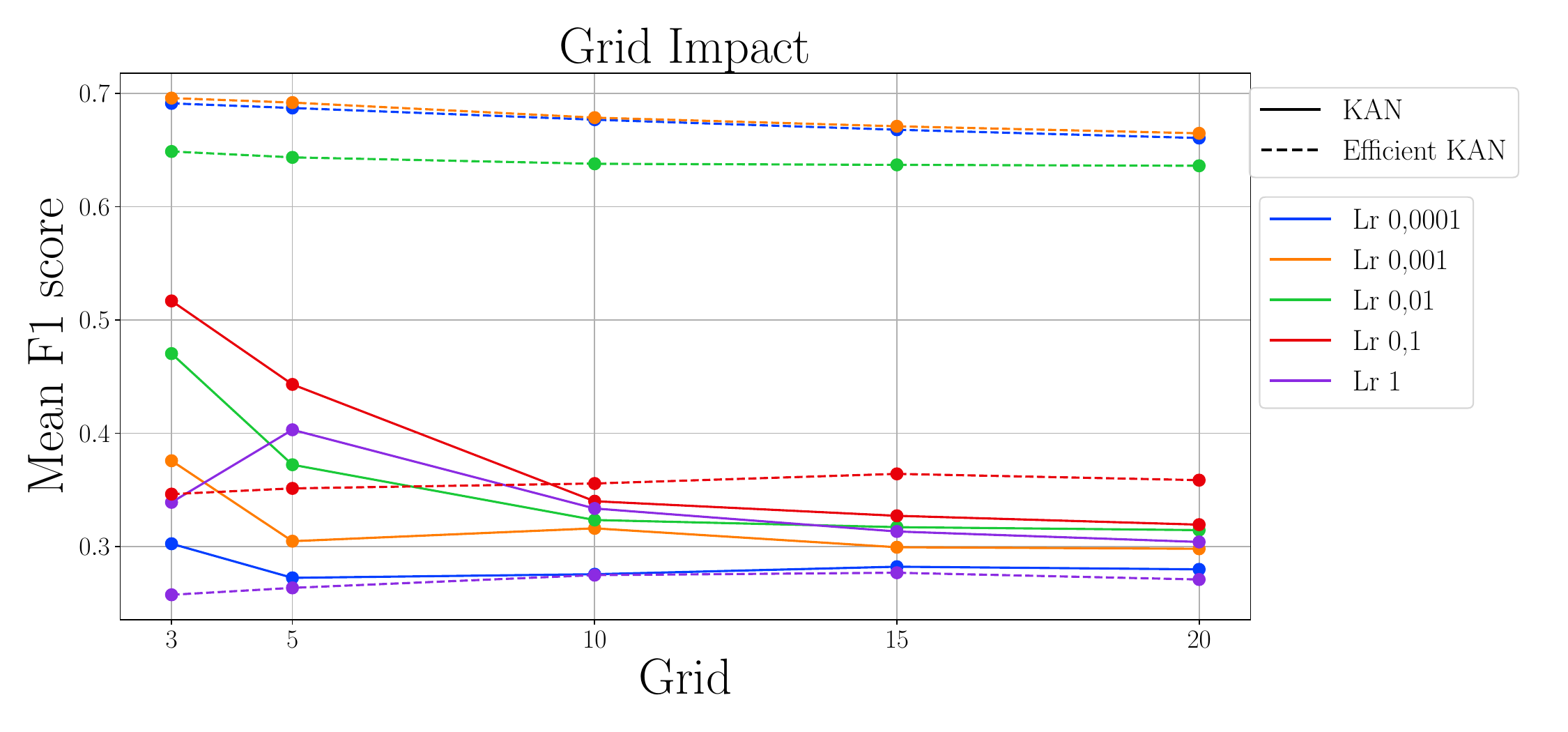}
    \caption{Grid impact on KAN and Efficient KAN models}
    \label{fig:grid_sensitivity}
\end{figure}
\begin{figure}[h]
    \centering
    \includegraphics[width=\linewidth]{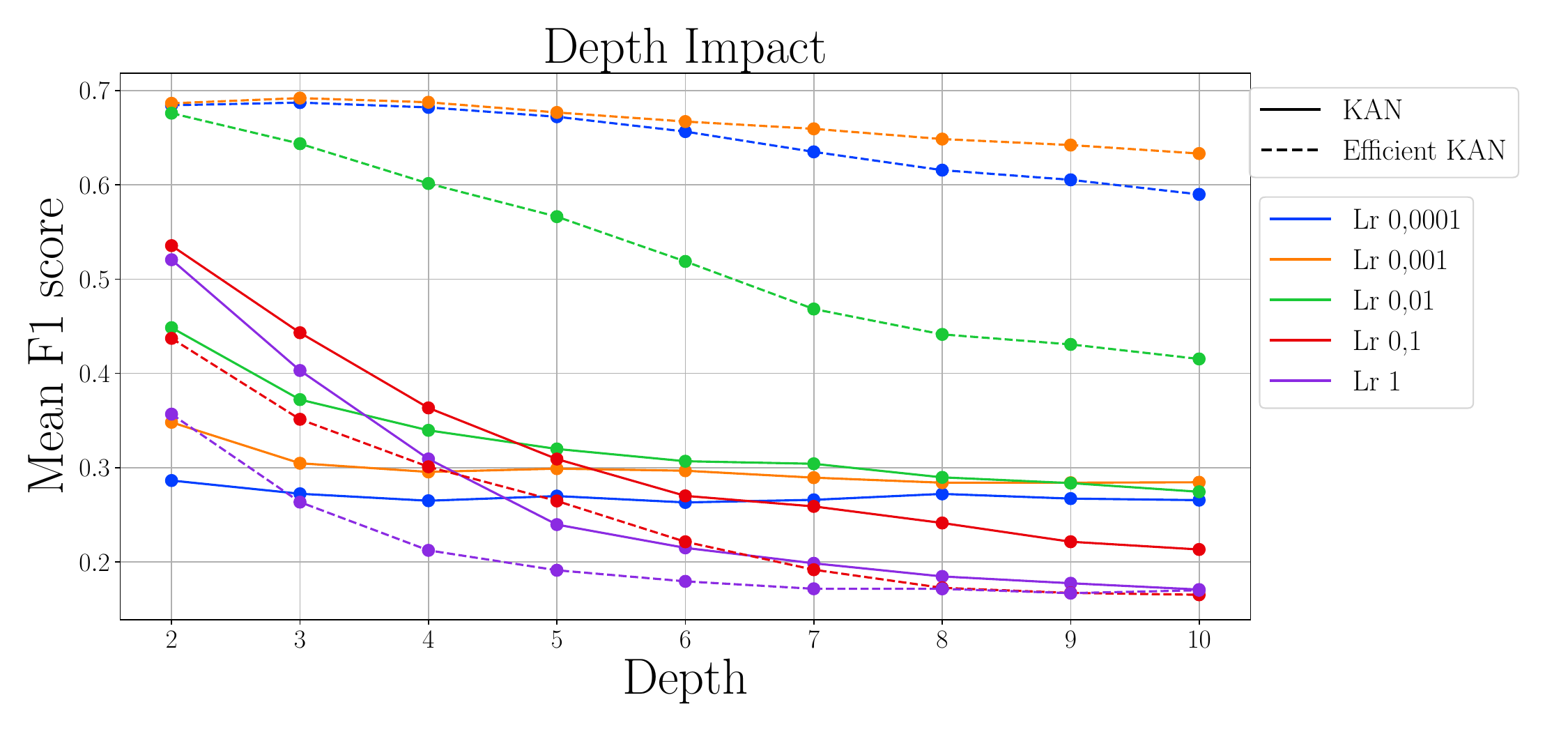}
    \caption{Depth impact on KAN and Efficient KAN models}
    \label{fig:depth_sensitivity}
\end{figure}
\begin{figure}[h]
    \centering
    \includegraphics[width=\linewidth]{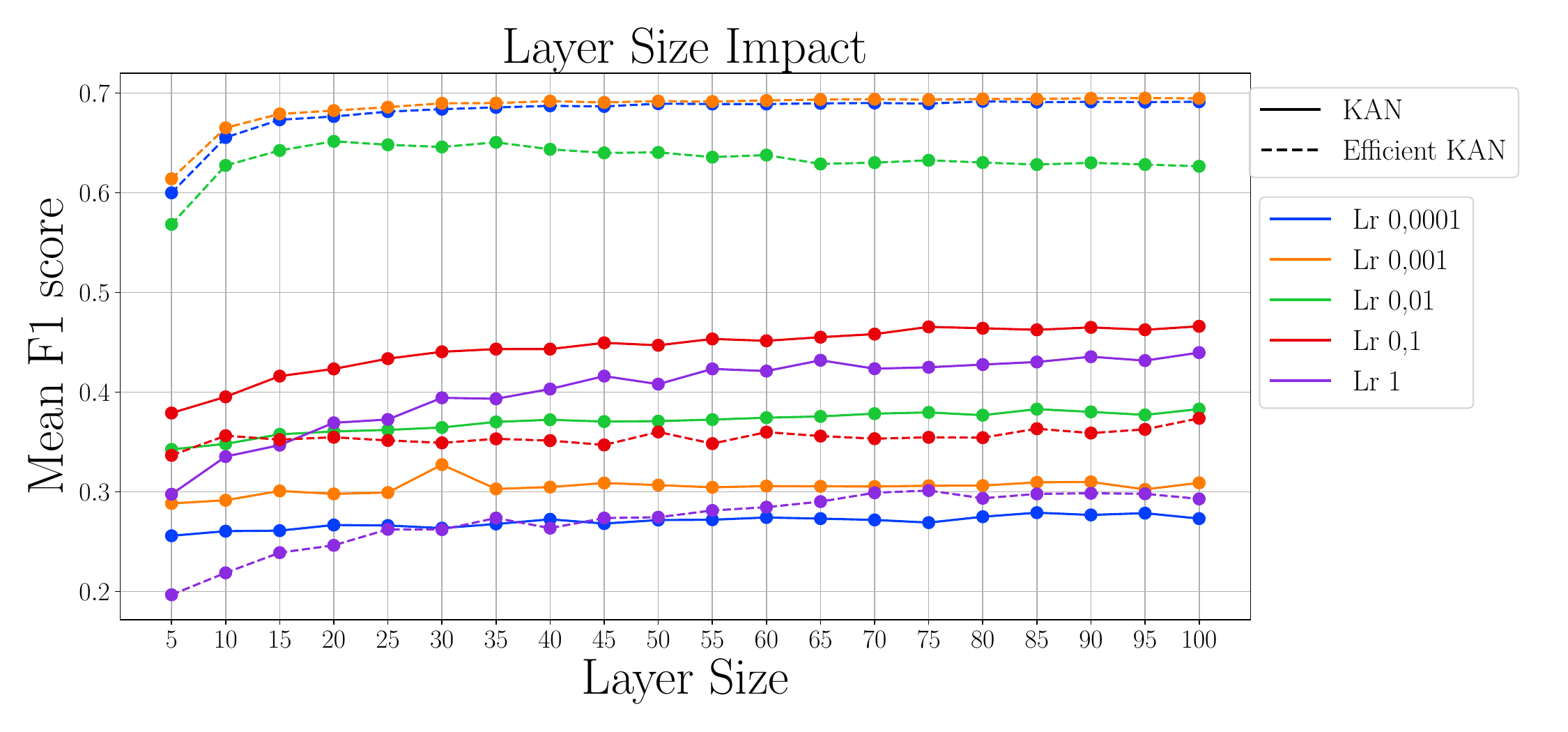}
    \caption{Layer size impact on KAN and Efficient KAN models}
    \label{fig:node_sensitivity}
\end{figure}

\subsection{Hyperparameter impact analysis}

In this section, we analyze the impact of key hyperparameters on KAN and Efficient KAN models, with Figs. \ref{fig:grid_sensitivity}, \ref{fig:depth_sensitivity} and \ref{fig:node_sensitivity} illustrating results for specific hyperparameter variations as detailed in Section \ref{sec:hyper_impact_methodology}. Each figure includes line plots for five different learning rates, where the same color represents the results with the same learning rate for both models, with KAN depicted with solid lines and Efficient KAN with dashed lines.

The impact of grid size \(G\) on model performance was evaluated across various learning rates, focusing on mean F1 score outcomes for both KAN and Efficient KAN. As it can be seen in Fig. \ref{fig:grid_sensitivity}, results show a general trend of decreasing F1 performance with increased grid size, which aligns with findings that larger grid configurations can lead to optimization challenges in KAN models \cite{dong2024kolmogorov}. This also suggests that while increasing grid size makes B-splines locally accurate, it potentially reduces the model's overall performance by making it globally inaccurate. While both models’ performances decrease with grid size, Efficient KAN maintains greater stability across grid configurations and learning rates, with lower learning rates proving more effective in preserving model performance.

At the lowest learning rates, Efficient KAN achieved the highest F1 scores, with \( lr=0.0001 \) and \( lr=0.001 \) yielding nearly identical and stable results across grid sizes. For these rates, the F1 score exhibited a slight, linear decrease from \(0.69\) at \(G=3\) to approximately \(0.66\) at \(G=20\), suggesting minimal sensitivity to grid variation. The learning rate \( lr=0.01 \) yielded slightly lower performance than \( lr =0.0001 \) and \( lr=0.001 \), following a similarly gradual decrease across grid sizes. In contrast, higher learning rates (\( lr=0.1 \) and \(\ lr=1 \)) resulted in significantly lower F1 scores, averaging around \(0.3\) across all grid sizes, showing limited grid sensitivity but almost 50\% reduced effectiveness relative to lower rates.

KAN, in contrast, exhibited an opposing learning rate trend, with performance improving as learning rates increased. All learning rates for KAN showed a similar pattern: a substantial drop in F1 score from \(G=3\) to \(G=10\), followed by a more gradual decline with increasing grid size. Notably, even KAN’s best configurations fell well below the best results of Efficient KAN, underscoring a clear performance gap. A deviation in KAN’s trend was seen at \( lr=1 \) for \(G=3\), where the F1 score was unexpectedly lower than at \(G=5\), unlike other rates where \(G=3\) consistently achieved the highest performance.

The effect of model depth reveals similar trends, as shown in Fig. \ref{fig:depth_sensitivity}. Efficient KAN achieved the highest performance at the smallest learning rates (\( lr=0.0001 \) and \( lr=0.001 \)), though differences between these rates became more pronounced with increasing depth. At a learning rate of \( lr=0.01 \), Efficient KAN’s performance declined more sharply with increasing depth, achieving a peak F1 score of \( 0.69 \) at depth 2, but dropping to \( 0.64 \) at depth 10. This pattern aligns with previous research indicating that higher learning rates can destabilize training by causing the model to overshoot optimal parameter values. In time-series tasks, lower learning rates in combination with Efficient KAN’s regularization mechanisms enable it to adjust gradually, allowing the model to learn nuanced temporal patterns without abrupt parameter updates, thereby supporting stable and high performance.

In contrast, KAN’s behavior across learning rates mirrored the patterns observed in the grid size analysis, with generally lower performance compared to Efficient KAN. For shallower depths (2–3), KAN achieved its best performance with higher learning rates; however, these rates led to a sharp performance drop as depth increased, with F1 scores falling below 0.2. Lower learning rates yielded a more gradual decline, stabilizing around 0.3 before slightly decreasing further. Regardless of the learning rate, increasing depth consistently resulted in lower performance. These results suggest that KAN performs best with smaller depths and benefits less from deeper configurations due to challenges in capturing complex time-series dependencies effectively at greater depths.

We further analyzed the effect of layer size, defined as the number of nodes in each layer, on model performance. Results are presented in Fig. \ref{fig:node_sensitivity}. For Efficient KAN, lower learning rates (\( lr = 0.0001 \) and \( lr = 0.001 \)) consistently yielded the best results, with \( lr = 0.01 \) performing slightly worse. Larger learning rates led to a significant performance drop, mirroring previous observations. Among layer sizes, configurations with 5 and 10 nodes per layer generally underperformed compared to larger sizes, which yielded stable F1 scores. These smaller configurations lack the complexity required to model diverse time-series data effectively, especially given the shortest series length of 15 data points. When layer sizes exceeded this threshold, Efficient KAN maintained consistent performance by avoiding overgeneralization and better capturing the data’s intricacies.

\begin{table*}[!htbp]
\centering
\begin{tabular}{|l|l|c|c|c|c|c|c|c|c|}
\hline
\multirow{2}{*}{\textbf{Model}} & \multirow{2}{*}{\textbf{Configuration}} & \multicolumn{2}{c|}{\textbf{Precision}} & \multicolumn{2}{c|}{\textbf{Recall}} & \multicolumn{2}{c|}{\textbf{F1 score}} & \multicolumn{2}{c|}{\textbf{Training time (seconds)}} \\
\cline{3-10}
& & \textbf{Mean} & \textbf{StdDev} & \textbf{Mean} & \textbf{StdDev} & \textbf{Mean} & \textbf{StdDev} & \textbf{Mean} & \textbf{StdDev}\\ 
\hline
KAN & [40], \( G = 5 \) & 0.58 & 0.19 & 0.56 & 0.21 & 0.54 & 0.22 & 6.36 & 2.8 \\  
Efficient KAN & [40, 40], \( G = 3 \) & 0.72 & 0.20 & 0.71 & 0.21 & 0.70 & 0.23 & 95.00 & 205.1 \\ 
MLP (3-depth) & [300, 300, 300] & 0.73 & 0.20 & 0.66 & 0.23 & 0.64 & 0.25  & 109.5 & 368.5\\
 InceptionTime & / & 0.84 & 0.17 & 0.83 & 0.18 & 0.84 & 0.18 & 4 006.60 & 5 049.82\\
Hive-Cote 2.0 & / & 0.87 & 0.13 & 0.85 & 0.16 & 0.85 & 0.17 & 14 700.32 & 57 228.95\\ 

\hline
\end{tabular}
\caption{Comparison of top-performing KAN models vs the reference MLP, InceptionTime and Hive-Cote2 SotA on 117 datasets in UCR archive.}
\label{tab:best_results}
\end{table*}

For KAN, the trend remained opposite: higher learning rates produced better F1 scores, with \( lr = 0.1 \) consistently yielding the highest performance. Unlike Efficient KAN, KAN’s performance gradually improved with increasing layer size, indicating that a larger number of nodes allowed it to capture more complex dependencies in the time-series data. A notable increase in F1 performance occurred with a layer configuration of 30 nodes per layer and a learning rate of \( lr = 0.001 \), highlighting this setup’s capacity to leverage both the model’s structure and time-series complexity. However, the worst-performing configuration for KAN was a high learning rate with very few nodes (5–20 per layer), which led to significant overgeneralization and poor performance. Smaller configurations likely struggle to represent complex patterns effectively, leading to overly simplistic models that underfit the data’s temporal dependencies.

In summary, the analyses of grid size, depth, and layer size reveal clear distinctions in how KAN and Efficient KAN respond to different hyperparameter configurations. Efficient KAN consistently performs best with smaller learning rates, showing stability across grid, depth, and layer size variations due to its inherent regularization mechanisms. This stability allows Efficient KAN to avoid overfitting and adapt effectively even with moderate layer sizes, particularly when layer sizes exceed the shortest time series length. In contrast, KAN’s optimal performance relies on higher learning rates and increased layer sizes, as these configurations enable it to capture more complex dependencies in the time-series data. KAN’s performance improves gradually with layer size and depth but is more sensitive to increases in grid size. Overall, Efficient KAN demonstrates a robust ability to maintain high performance with conservative hyperparameter settings, whereas KAN benefits more from expanded configurations to reach its optimal performance.

Out of all computed results across various hyperparameter configurations, we selected the best-performing KAN and Efficient KAN models. Their results are presented in Table \ref{tab:best_results}. The table is structured in a similar manner as Table \ref{tab:model_performance}, along with additional column showing mean and standard deviation of average training time in seconds.

\begin{figure}[!htbp]
    \centering
    \includegraphics[width=\linewidth]{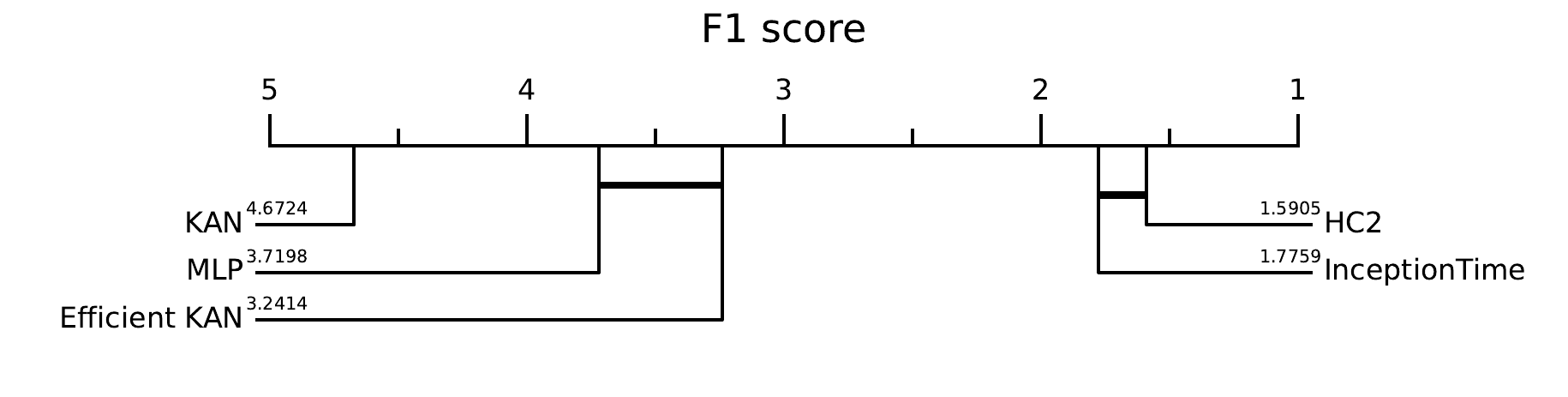}
    \caption{ Critical diagram of F1 score for five models}
    \label{fig:cd_graph-f1-new}
\end{figure}

For comparison, we also include results from the MLP model discussed in Section \ref{sec:class_performance},  InceptionTime~\cite{ismail2020inceptiontime}  baseline and Hive-Cote 2.0 as a state-of-the-art benchmark. Efficient KAN demonstrates comparable performance to MLP, with higher F1 scores and significantly faster training times. Although Hive-Cote 2.0 achieves the highest overall performance, Efficient KAN provides a balanced tradeoff with substantially reduced training time, making it a practical alternative for time-sensitive applications. This is further illustrated in the critical difference diagrams in Fig. \ref{fig:cd_graph-f1-new}, where Efficient KAN ranks above MLP but still below Hive-Cote 2.0  and InceptionTime, reflecting its competitive accuracy while maintaining efficiency. In contrast, KAN ranks lowest, showing a more significant deviation in performance.
\begin{figure}[thp]
  \centering

  \begin{subfigure}[b]{0.39\textwidth}
    \includegraphics[width=\textwidth]{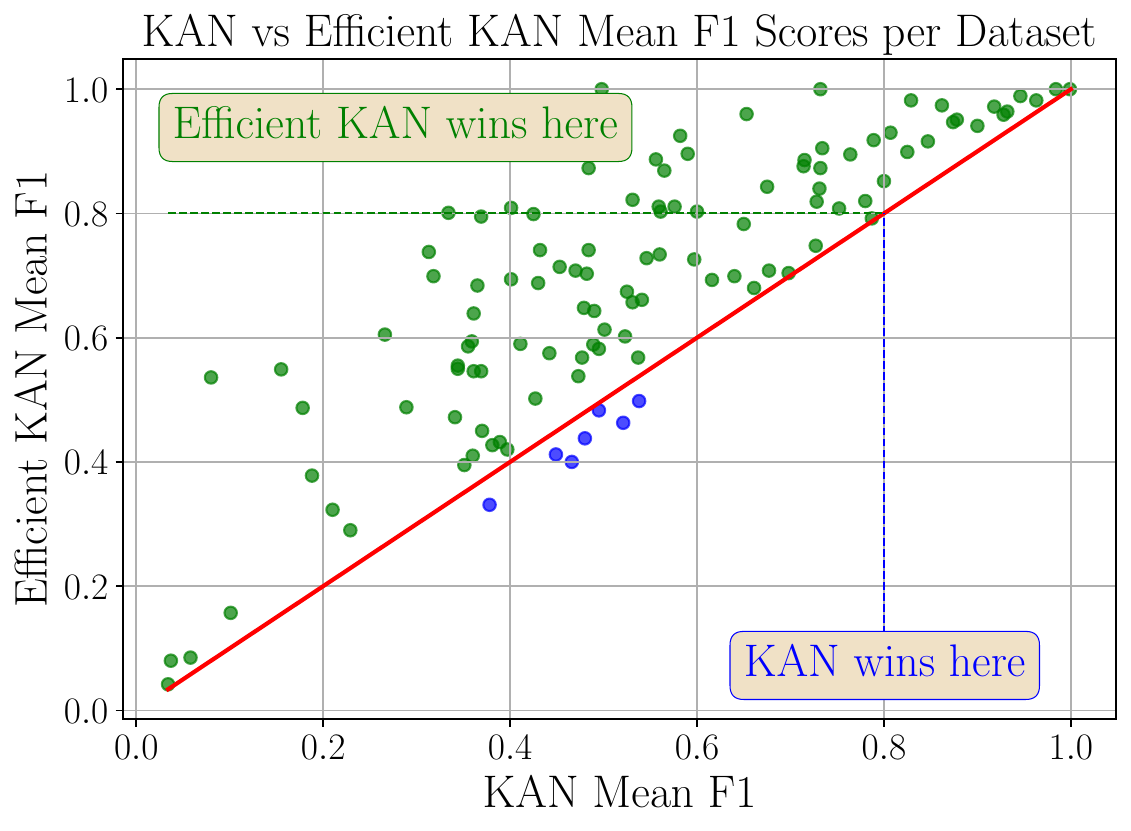}
    \caption{Efficient KAN vs KAN}
    \label{fig:kan_vs_effkan}
  \end{subfigure}

  \vspace{1.5em} %

  \begin{subfigure}[b]{0.39\textwidth}
    \includegraphics[width=\textwidth]{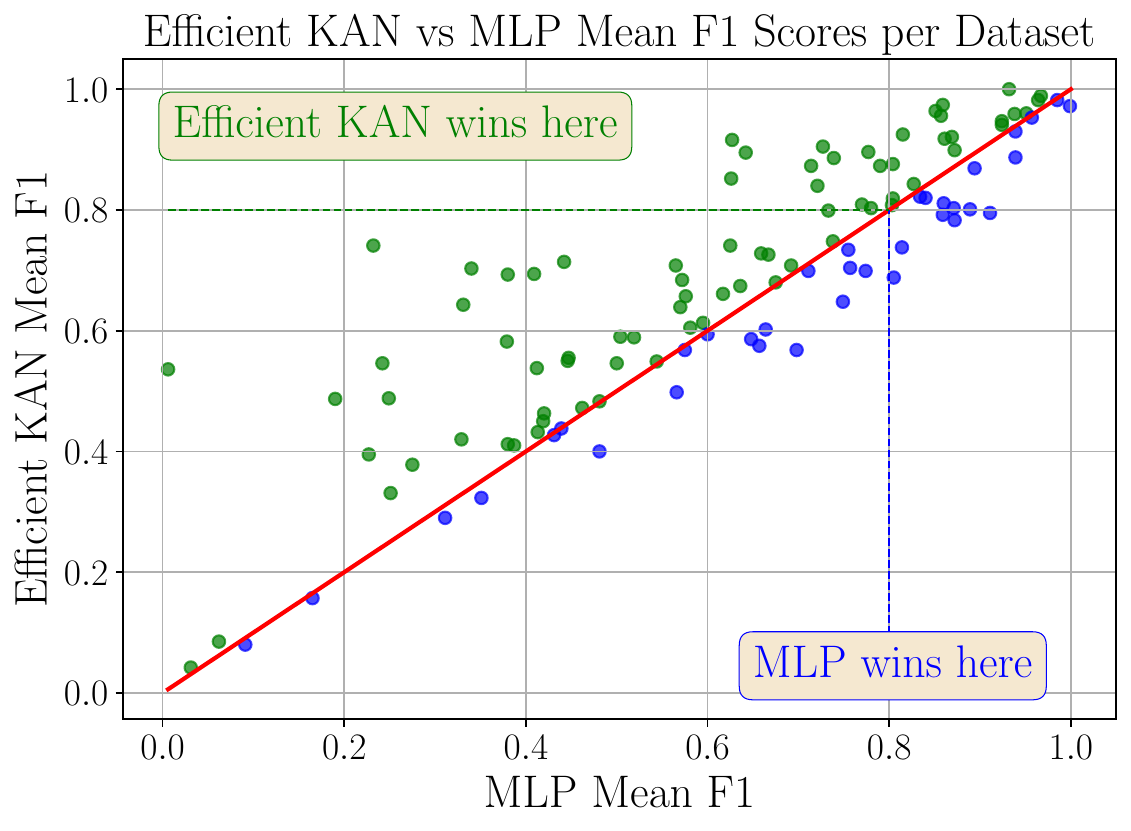}
    \caption{Efficient KAN vs MLP}
    \label{fig:effkan_vs_mlp}
  \end{subfigure}

  \vspace{1.5em} %

  \begin{subfigure}[b]{0.39\textwidth}
    \includegraphics[width=\textwidth]{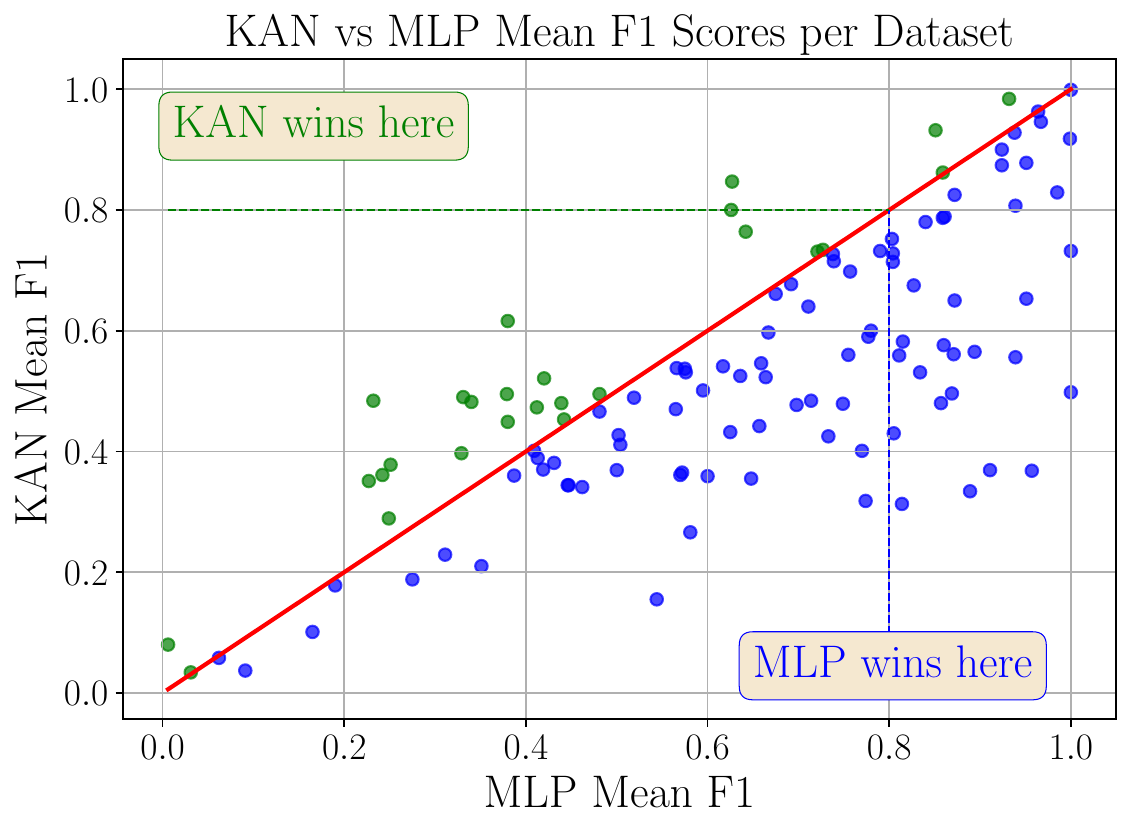}
    \caption{KAN vs MLP}
    \label{fig:kan_vs_mlp}
  \end{subfigure}

  \caption{ Scatter plots comparing mean F1 scores of different model pairs.}
  \label{fig:scatter_comparisons}
\end{figure}
 To provide a more detailed insight into comparison between the three best developed models, we also include three scatter-plots in Fig.~\ref{fig:scatter_comparisons} , that visualize pairwise comparisons between them: KAN vs. MLP, Efficient KAN vs. MLP, and KAN vs. Efficient KAN. In each scatterplot, the mean F1-score across 5 different runs for each dataset obtained by one model is plotted on the x-axis, while the mean F1-score of the other model is plotted on the y-axis. These plots provide an intuitive visual comparison of relative performance across datasets. Points above the diagonal line indicate datasets where the model on the y-axis outperforms the one on the x-axis.

 In Figure~\ref{fig:kan_vs_effkan} , Efficient KAN is compared to the original KAN. The majority of points lie above the diagonal, indicating that Efficient KAN achieves higher F1-scores than the original KAN on most datasets. Similarly, scatter plot in Figure~\ref{fig:effkan_vs_mlp}  confirms that Efficient KAN generally outperforms MLP. In Figure~\ref{fig:kan_vs_mlp} , original KAN is compared to MLP. While MLP outperforms KAN on a number of datasets, the distinction is less prominent compared to the previous two comparisons.

\subsection{Interpretability analysis}

In this section, we evaluate the interpretability of the KAN model and compare it with the interpretability of MLP, following the methodology described in Section~\ref{sec:interpret_methodology}. The interpretability of the two models is demonstrated using the SmoothSubspace dataset from the UCR archive shown in Fig.~\ref{fig:ts_classes}. We compare the interpretability of the KAN model with the best MLP configuration (see Table~\ref{tab:best_results}), which achieved an F1 score of almost 0.88 on the used dataset, which is similar to the fine-tuned KAN model utilised to demonstrate interpretability described in Section~ \ref{sec:interpret_methodology}.

\begin{figure}[h!]
    \centering
    \includegraphics[width=\linewidth]{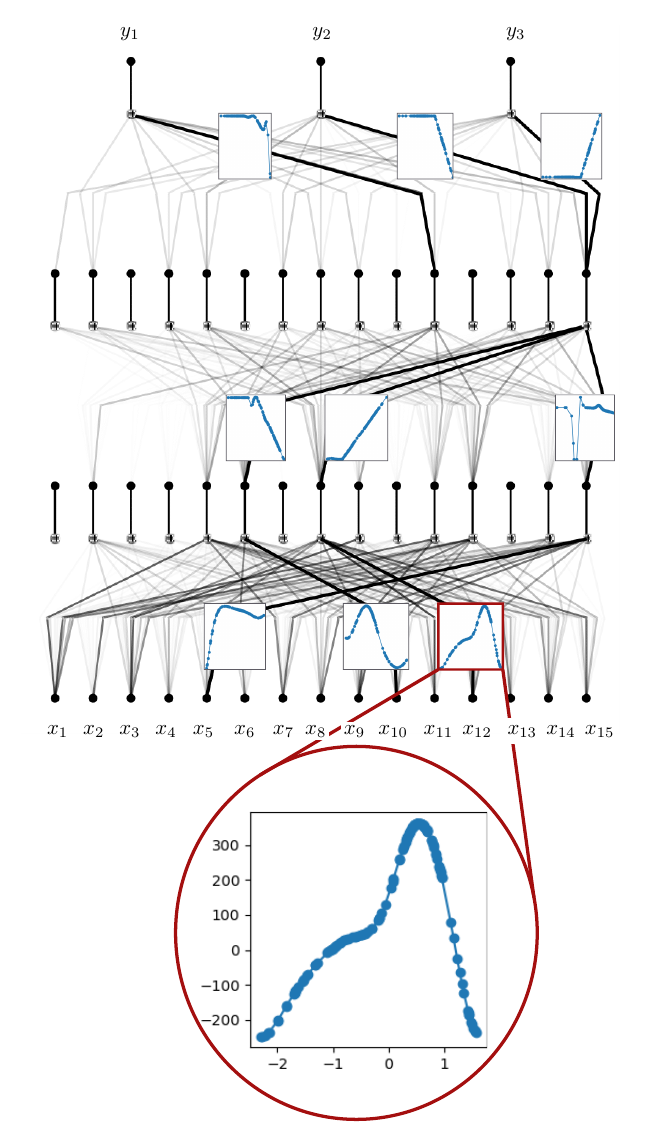}
    \caption{KAN model with edge thickness indicating connection importance and B-spline plots showing learned functions on edges.}
    \label{fig:model}
\end{figure}

 Unlike traditional neural networks such as MLPs, which require post-hoc techniques like SHAP to gain interpretability, KANs offer interpretability by design through their composition graph. Fig.~\ref{fig:model} provides a visualization of the structure of the KAN model, where the thickness of the edges represents the importance of the connections based on the edge scores according to Section~\ref{sec:interpret_methodology}. The corresponding B-splines for these edges show which function is applied to the input features and provide information on how the input features are modified by the architecture. The x-axis of the enhanced B-spline in Fig.~\ref{fig:model} represents the range of input features, while the y-axis indicates the corresponding outputs of the spline function applied to these inputs.

Looking at the KAN graph in Fig.~\ref{fig:model}, it can be seen that features $x_5$, $x_{10}$, and $x_{12}$ emerge as the most influential for the models decision, based on the edge weights where thicker lines represent higher edge weight. Looking at Fig.~\ref{fig:ts_classes} that input features $x_5$, $x_{10}$ are effective in differentiating class $y_2$ from classes $y_1$ and $y_3$, while feature $x_{12}$ further distinguishes between classes $y_1$ and $y_3$. Additionally, looking at class $y_3$ in Fig.~\ref{fig:ts_classes}, we can see that majority of the values within the dataset are whithin $[0, 1]$ for input feature $x_{12}$. Looking at the learned B-spline corresponding to this feature in Fig.~\ref{fig:model} we can see that when $x_{12}$ is between 0 and 1, the spline increases the values up to 300, while the values outside of this range decrease to values between 0 and -200, diminishing their contribution. Similar observations can also be made for the input features $x_5$ and $x_{10}$.  In contrast, looking at feature $x_9$, we can see that it provides limited relevance for classification indicated by its thinner connection edges in Fig.~\ref{fig:model} . As seen in Fig.~\ref{fig:ts_classes},  it helps separating class $y_2$, but offers little distinction between classes $y_1$ and $y_3$, where values largely overlap. This aligns with the low model sensitivity reflected in the KAN graph.

\begin{figure*}[h!]
\centering
\includegraphics[width=\textwidth]{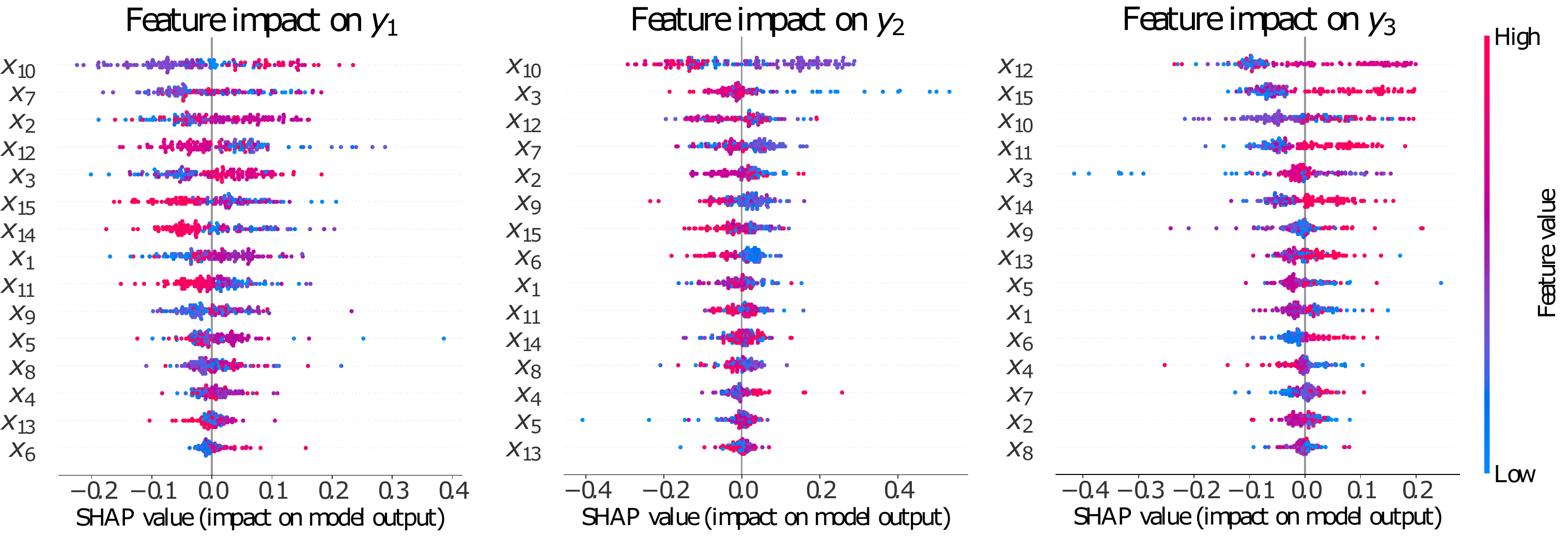}
\caption{SHAP summary plots showing feature impacts in KAN model}
\label{fig:shap_kan}
\end{figure*}

\begin{figure*}[h!]
\centering
\includegraphics[width=\textwidth]{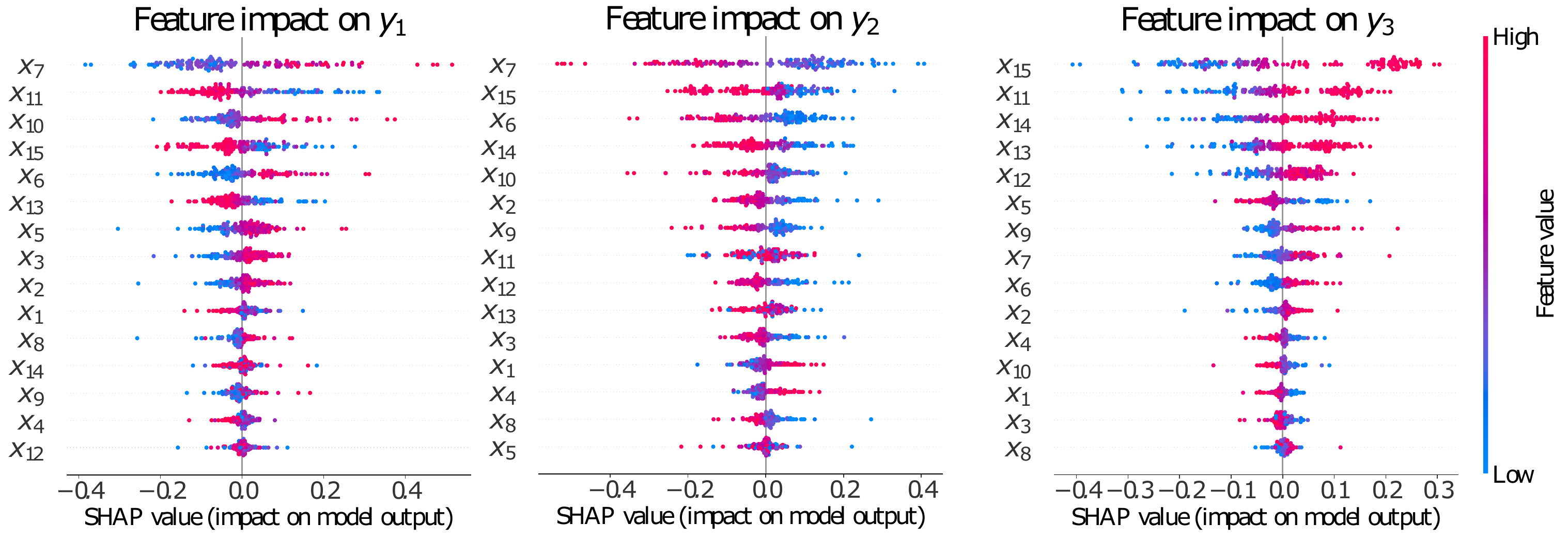}
\caption{SHAP summary plots showing feature impacts in MLP model}
\label{fig:shap_mlp}
\end{figure*}

To  validate and further quantify feature contributions and also compare feature importance of KAN to MLP, we apply SHAP analysis as described in Section~\ref{sec:interpret_methodology}.  This allows us to validate whether the influential features identified through KAN’s structure and learned B-splines in Fig.~\ref{fig:model}  are also reflected consistently by the more established SHAP explanations. Figs.~\ref{fig:shap_kan} and \ref{fig:shap_mlp} display the feature importance for each class of the SmoothSubspace dataset for both KAN and MLP, sorted from the most important to the least important feature for each predicted class. Each figure is divided into three sections, corresponding to the three output classes. In the plots, each feature’s contribution is represented by its SHAP value, which indicates whether it positively contributes to the prediction or negatively. Feature values are color-coded, with red indicating high values and blue low values. As it can be seen in Fig.~\ref{fig:ts_classes}, $2$ represents the highest value, while $-2$ represents the lowest value. Middle-range values between $-1$ to $1$, are shown as purple dots. Purple dots correspond to the flat parts of the series, where consecutive values show minimal variation, while red and blue dots are associated with the highly fluctuating sections of the series. 

Examining the SHAP plot for the KAN model in Fig.~\ref{fig:shap_kan}, it can be seen that the top contributors, features $x_{10}$ and $x_{12}$ consistently stand out, aligning with KAN’s interpretability findings from Fig.~\ref{fig:model}. Interestingly, compared to the feature importance from Fig.~\ref{fig:model}, feature $x_5$ is not present among the top contributors according to SHAP. This contrast shows how different interpretability methods can yield complementary insights, each capturing unique aspects of feature importance.

 For the MLP model, which lacks inherent interpretability, SHAP is essential. Fig.~\ref{fig:shap_mlp}  presents SHAP-based feature importances for MLP. As it can be seen, feature $x_7$ is the most important in distinguishing classes $y_1$ and $y_2$, compared to $x_{10}$ for KAN model. However, for both MLP and KAN the $x_{10}$ and $x_7$ are shown as top 4 contributing features in distinguishing between the classes, indicating that both MLP and KAN model in general consider similar features as key to distinguish between the three classes. 

 However, while both models highlight similar features overall, it is evident that for KAN the SHAP color coding aligns consistently with the structure of the time series, whereas in the MLP the color mapping appears more scattered and less coherent with the underlying sequential patterns, which makes its feature contributions harder to interpret in relation to the original series.

In summary, KAN provides interpretability inherently through its graph structure and learned splines. SHAP is employed primarily to validate and complement this built-in transparency. In contrast, MLP requires SHAP or similar tools to expose any interpretability, underscoring the advantage of KANs in applications where understanding model decisions is critical.

\section{Conclusion}
\label{sec:conclusion}

The aim of this study was to conduct a comprehensive and robust exploration of the KAN architecture for time series classification. To achieve that, we first investigated the transferability of reference Kolmogorov-Arnold Networks (KANs) from regression to classification tasks , running a large number of models across 117 diverse datasets from the UCR Benchmark. We conducted hyperparameter search on two KAN implementations in view of finding the best architecture for the task at hand and concluded with an interpretability analysis. 
Our findings show that the Efficient KAN implementation has a significant performance improvement over the original KAN and is also superior to MLPs by 6 percentage points in F1 score on the reference architectures. The hyperparameter analysis revealed that Efficient KAN consistently outperforms KAN across variations in grid size, depth, and layer size, showcasing its robustness and adaptability to diverse datasets. Efficient KAN proved more stable than KAN across grid sizes, depths, and layer configurations, particularly with lower learning rates. The interpretability of KANs was confirmed through SHAP analysis and B-spline visualization, highlighting their potential for transparent decision-making. In contrast, MLPs, though competitive in performance, exhibited less interpretability. Future work could explore hybrid architectures and further optimization to enhance KAN's application scope.

\section*{Funding}
This work was supported by the Slovenian Research Agency (Javna Agencija za Raziskovalno Dejavnost RS) under grant P2-0016. A preprint has previously been published \cite{baravsin2024exploring}.

Conflict of Interest: The authors declare that they have no conflict of interest.

\bibliographystyle{IEEEtran}
\bibliography{./references}

\onecolumn
\appendix

In this section, we present a summary table of the average F1-scores for the best-performing models identified during our experiments. Specifically, this includes the Kolmogorov–Arnold Network (KAN) with configuration [40], G = 5 and learning rate 0.1, the Efficient KAN variant with configuration [40,40], G = 3 and learning rate 0.001, and a baseline Multi-Layer Perceptron (MLP) with three hidden layers of size 300 and a learning rate of 0.001. The results are reported as mean F1-scores averaged across five random seeds to ensure robustness and comparability. The table is structured as follows: The first column lists the dataset names, followed by the size of the training and test subsets. The next column shows the length of the time series (input) and the number of classes (output), which correspond to the input and output layer sizes of the models. The remaining columns display the average F1-scores and standard deviations for each model configuration. 

{\small
\begin{longtable}{lccccc}
\toprule
Name & Train/Test & Length/Class & KAN & Efficient KAN & MLP \\
\midrule
\endfirsthead
\toprule
Name & Train/Test & Length/Class & KAN & Efficient KAN & MLP \\
\midrule
\endhead
\midrule
\multicolumn{6}{r}{Continued on next page} \\
\midrule
\endfoot
\bottomrule
\endlastfoot
Adiac & 390 / 391 & 176 / 37 & 0.401 $\pm$ 0.051 & 0.694 $\pm$ 0.019 & 0.409 $\pm$ 0.026 \\
ArrowHead & 36 / 175 & 251 / 3 & 0.430 $\pm$ 0.039 & 0.688 $\pm$ 0.005 & 0.805 $\pm$ 0.013 \\
Beef & 30 / 30 & 470 / 5 & 0.334 $\pm$ 0.092 & 0.801 $\pm$ 0.000 & 0.889 $\pm$ 0.015 \\
BeetleFly & 20 / 20 & 512 / 2 & 0.565 $\pm$ 0.065 & 0.869 $\pm$ 0.028 & 0.894 $\pm$ 0.001 \\
BirdChicken & 20 / 20 & 512 / 2 & 0.560 $\pm$ 0.061 & 0.734 $\pm$ 0.000 & 0.755 $\pm$ 0.051 \\
Car & 60 / 60 & 577 / 4 & 0.600 $\pm$ 0.076 & 0.803 $\pm$ 0.034 & 0.780 $\pm$ 0.016 \\
CBF & 30 / 900 & 128 / 3 & 0.480 $\pm$ 0.063 & 0.956 $\pm$ 0.006 & 0.857 $\pm$ 0.002 \\
ChlorineConcentration & 467 / 3840 & 166 / 3 & 0.484 $\pm$ 0.019 & 0.741 $\pm$ 0.007 & 0.232 $\pm$ 0.000 \\
CinCECGTorso & 40 / 1380 & 1639 / 4 & 0.425 $\pm$ 0.075 & 0.799 $\pm$ 0.010 & 0.733 $\pm$ 0.009 \\
Coffee & 28 / 28 & 286 / 2 & 0.498 $\pm$ 0.076 & 1.000 $\pm$ 0.000 & 1.000 $\pm$ 0.000 \\
Computers & 250 / 250 & 720 / 2 & 0.538 $\pm$ 0.023 & 0.498 $\pm$ 0.007 & 0.566 $\pm$ 0.030 \\
CricketX & 390 / 390 & 300 / 12 & 0.344 $\pm$ 0.009 & 0.550 $\pm$ 0.012 & 0.446 $\pm$ 0.006 \\
CricketY & 390 / 390 & 300 / 12 & 0.369 $\pm$ 0.016 & 0.546 $\pm$ 0.010 & 0.500 $\pm$ 0.012 \\
CricketZ & 390 / 390 & 300 / 12 & 0.344 $\pm$ 0.022 & 0.555 $\pm$ 0.017 & 0.447 $\pm$ 0.025 \\
DiatomSizeReduction & 16 / 306 & 345 / 4 & 0.401 $\pm$ 0.047 & 0.809 $\pm$ 0.020 & 0.770 $\pm$ 0.013 \\
DistalPhalanxOutlineAgeGroup & 400 / 139 & 80 / 3 & 0.489 $\pm$ 0.049 & 0.589 $\pm$ 0.007 & 0.519 $\pm$ 0.016 \\
DistalPhalanxOutlineCorrect & 600 / 276 & 80 / 2 & 0.677 $\pm$ 0.043 & 0.708 $\pm$ 0.004 & 0.692 $\pm$ 0.014 \\
DistalPhalanxTW & 400 / 139 & 80 / 6 & 0.397 $\pm$ 0.026 & 0.420 $\pm$ 0.011 & 0.329 $\pm$ 0.010 \\
Earthquakes & 322 / 139 & 512 / 2 & 0.495 $\pm$ 0.028 & 0.483 $\pm$ 0.025 & 0.481 $\pm$ 0.011 \\
ECG200 & 100 / 100 & 96 / 2 & 0.714 $\pm$ 0.075 & 0.876 $\pm$ 0.010 & 0.804 $\pm$ 0.022 \\
ECG5000 & 500 / 4500 & 140 / 5 & 0.495 $\pm$ 0.025 & 0.582 $\pm$ 0.007 & 0.379 $\pm$ 0.001 \\
ECGFiveDays & 23 / 861 & 136 / 2 & 0.556 $\pm$ 0.053 & 0.887 $\pm$ 0.005 & 0.939 $\pm$ 0.006 \\
ElectricDevices & 8926 / 7711 & 96 / 7 & 0.289 $\pm$ 0.032 & 0.488 $\pm$ 0.007 & 0.249 $\pm$ 0.016 \\
FaceAll & 560 / 1690 & 131 / 14 & 0.546 $\pm$ 0.042 & 0.728 $\pm$ 0.012 & 0.659 $\pm$ 0.007 \\
FaceFour & 24 / 88 & 350 / 4 & 0.318 $\pm$ 0.027 & 0.699 $\pm$ 0.011 & 0.774 $\pm$ 0.021 \\
FacesUCR & 200 / 2050 & 131 / 14 & 0.432 $\pm$ 0.030 & 0.741 $\pm$ 0.008 & 0.625 $\pm$ 0.012 \\
FiftyWords & 450 / 455 & 270 / 50 & 0.178 $\pm$ 0.019 & 0.487 $\pm$ 0.005 & 0.190 $\pm$ 0.015 \\
Fish & 175 / 175 & 463 / 7 & 0.675 $\pm$ 0.035 & 0.843 $\pm$ 0.006 & 0.827 $\pm$ 0.014 \\
FordA & 3601 / 1320 & 500 / 2 & 0.482 $\pm$ 0.063 & 0.703 $\pm$ 0.007 & 0.340 $\pm$ 0.000 \\
FordB & 3636 / 810 & 500 / 2 & 0.490 $\pm$ 0.022 & 0.643 $\pm$ 0.012 & 0.331 $\pm$ 0.000 \\
GunPoint & 50 / 150 & 150 / 2 & 0.734 $\pm$ 0.081 & 0.905 $\pm$ 0.005 & 0.727 $\pm$ 0.041 \\
Ham & 109 / 105 & 431 / 2 & 0.640 $\pm$ 0.050 & 0.699 $\pm$ 0.007 & 0.711 $\pm$ 0.012 \\
HandOutlines & 1000 / 370 & 2709 / 2 & 0.787 $\pm$ 0.048 & 0.792 $\pm$ 0.007 & 0.859 $\pm$ 0.016 \\
Haptics & 155 / 308 & 1092 / 5 & 0.389 $\pm$ 0.026 & 0.432 $\pm$ 0.007 & 0.413 $\pm$ 0.017 \\
Herring & 64 / 64 & 512 / 2 & 0.477 $\pm$ 0.070 & 0.568 $\pm$ 0.028 & 0.698 $\pm$ 0.035 \\
InlineSkate & 100 / 550 & 1882 / 7 & 0.229 $\pm$ 0.015 & 0.290 $\pm$ 0.009 & 0.311 $\pm$ 0.010 \\
InsectWingbeatSound & 220 / 1980 & 256 / 11 & 0.359 $\pm$ 0.047 & 0.594 $\pm$ 0.004 & 0.600 $\pm$ 0.015 \\
ItalyPowerDemand & 67 / 1029 & 24 / 2 & 0.878 $\pm$ 0.020 & 0.951 $\pm$ 0.002 & 0.951 $\pm$ 0.005 \\
LargeKitchenAppliances & 375 / 375 & 720 / 3 & 0.480 $\pm$ 0.034 & 0.438 $\pm$ 0.013 & 0.439 $\pm$ 0.011 \\
Lightning2 & 60 / 61 & 637 / 2 & 0.661 $\pm$ 0.088 & 0.680 $\pm$ 0.026 & 0.675 $\pm$ 0.037 \\
Lightning7 & 70 / 73 & 319 / 7 & 0.361 $\pm$ 0.057 & 0.639 $\pm$ 0.017 & 0.570 $\pm$ 0.028 \\
Mallat & 55 / 2345 & 1024 / 8 & 0.496 $\pm$ 0.053 & 0.921 $\pm$ 0.008 & 0.869 $\pm$ 0.041 \\
Meat & 60 / 60 & 448 / 3 & 0.807 $\pm$ 0.125 & 0.930 $\pm$ 0.006 & 0.939 $\pm$ 0.009 \\
MedicalImages & 381 / 760 & 99 / 10 & 0.361 $\pm$ 0.039 & 0.546 $\pm$ 0.015 & 0.242 $\pm$ 0.028 \\
MiddlePhalanxOutlineAgeGroup & 400 / 154 & 80 / 3 & 0.351 $\pm$ 0.043 & 0.395 $\pm$ 0.024 & 0.227 $\pm$ 0.043 \\
MiddlePhalanxOutlineCorrect & 600 / 291 & 80 / 2 & 0.521 $\pm$ 0.043 & 0.463 $\pm$ 0.013 & 0.420 $\pm$ 0.027 \\
MiddlePhalanxTW & 399 / 154 & 80 / 6 & 0.378 $\pm$ 0.026 & 0.331 $\pm$ 0.010 & 0.251 $\pm$ 0.002 \\
MoteStrain & 20 / 1252 & 84 / 2 & 0.576 $\pm$ 0.088 & 0.811 $\pm$ 0.013 & 0.860 $\pm$ 0.002 \\
NonInvasiveFetalECGThorax1 & 1800 / 1965 & 750 / 42 & 0.590 $\pm$ 0.038 & 0.896 $\pm$ 0.005 & 0.777 $\pm$ 0.009 \\
NonInvasiveFetalECGThorax2 & 1800 / 1965 & 750 / 42 & 0.582 $\pm$ 0.049 & 0.925 $\pm$ 0.002 & 0.815 $\pm$ 0.019 \\
OliveOil & 30 / 30 & 570 / 4 & 0.531 $\pm$ 0.047 & 0.822 $\pm$ 0.033 & 0.834 $\pm$ 0.028 \\
OSULeaf & 200 / 242 & 427 / 6 & 0.427 $\pm$ 0.023 & 0.502 $\pm$ 0.020 & 0.502 $\pm$ 0.010 \\
PhalangesOutlinesCorrect & 1800 / 858 & 80 / 2 & 0.616 $\pm$ 0.031 & 0.693 $\pm$ 0.014 & 0.380 $\pm$ 0.000 \\
Phoneme & 214 / 1896 & 1024 / 39 & 0.034 $\pm$ 0.004 & 0.042 $\pm$ 0.001 & 0.031 $\pm$ 0.003 \\
Plane & 105 / 105 & 144 / 7 & 0.829 $\pm$ 0.060 & 0.982 $\pm$ 0.000 & 0.985 $\pm$ 0.005 \\
ProximalPhalanxOutlineAgeGroup & 400 / 205 & 80 / 3 & 0.698 $\pm$ 0.035 & 0.704 $\pm$ 0.008 & 0.757 $\pm$ 0.016 \\
ProximalPhalanxOutlineCorrect & 600 / 291 & 80 / 2 & 0.780 $\pm$ 0.026 & 0.820 $\pm$ 0.009 & 0.840 $\pm$ 0.018 \\
ProximalPhalanxTW & 400 / 205 & 80 / 6 & 0.473 $\pm$ 0.016 & 0.538 $\pm$ 0.013 & 0.412 $\pm$ 0.012 \\
RefrigerationDevices & 375 / 375 & 720 / 3 & 0.449 $\pm$ 0.030 & 0.412 $\pm$ 0.009 & 0.380 $\pm$ 0.009 \\
ScreenType & 375 / 375 & 720 / 3 & 0.360 $\pm$ 0.021 & 0.410 $\pm$ 0.014 & 0.387 $\pm$ 0.013 \\
ShapeletSim & 20 / 180 & 500 / 2 & 0.466 $\pm$ 0.053 & 0.400 $\pm$ 0.043 & 0.481 $\pm$ 0.010 \\
ShapesAll & 600 / 600 & 512 / 60 & 0.365 $\pm$ 0.060 & 0.684 $\pm$ 0.007 & 0.572 $\pm$ 0.015 \\
SmallKitchenAppliances & 375 / 375 & 720 / 3 & 0.341 $\pm$ 0.049 & 0.472 $\pm$ 0.012 & 0.462 $\pm$ 0.012 \\
SonyAIBORobotSurface1 & 20 / 601 & 70 / 2 & 0.523 $\pm$ 0.093 & 0.602 $\pm$ 0.008 & 0.664 $\pm$ 0.023 \\
SonyAIBORobotSurface2 & 27 / 953 & 65 / 2 & 0.559 $\pm$ 0.056 & 0.811 $\pm$ 0.003 & 0.811 $\pm$ 0.002 \\
StarLightCurves & 1000 / 8236 & 1024 / 3 & 0.847 $\pm$ 0.024 & 0.916 $\pm$ 0.003 & 0.627 $\pm$ 0.001 \\
Strawberry & 613 / 370 & 235 / 2 & 0.928 $\pm$ 0.018 & 0.959 $\pm$ 0.002 & 0.938 $\pm$ 0.005 \\
SwedishLeaf & 500 / 625 & 128 / 15 & 0.715 $\pm$ 0.037 & 0.886 $\pm$ 0.006 & 0.739 $\pm$ 0.023 \\
Symbols & 25 / 995 & 398 / 6 & 0.313 $\pm$ 0.091 & 0.738 $\pm$ 0.007 & 0.814 $\pm$ 0.019 \\
SyntheticControl & 300 / 300 & 60 / 6 & 0.900 $\pm$ 0.016 & 0.941 $\pm$ 0.006 & 0.924 $\pm$ 0.012 \\
ToeSegmentation1 & 40 / 228 & 277 / 2 & 0.501 $\pm$ 0.034 & 0.613 $\pm$ 0.012 & 0.595 $\pm$ 0.006 \\
ToeSegmentation2 & 36 / 130 & 343 / 2 & 0.442 $\pm$ 0.033 & 0.575 $\pm$ 0.022 & 0.657 $\pm$ 0.016 \\
Trace & 100 / 100 & 275 / 4 & 0.800 $\pm$ 0.038 & 0.852 $\pm$ 0.014 & 0.626 $\pm$ 0.016 \\
TwoLeadECG & 23 / 1139 & 82 / 2 & 0.561 $\pm$ 0.070 & 0.803 $\pm$ 0.007 & 0.871 $\pm$ 0.026 \\
TwoPatterns & 1000 / 4000 & 128 / 4 & 0.789 $\pm$ 0.041 & 0.918 $\pm$ 0.009 & 0.861 $\pm$ 0.011 \\
UWaveGestureLibraryAll & 896 / 3582 & 945 / 8 & 0.874 $\pm$ 0.024 & 0.947 $\pm$ 0.003 & 0.924 $\pm$ 0.008 \\
UWaveGestureLibraryX & 896 / 3582 & 315 / 8 & 0.597 $\pm$ 0.017 & 0.726 $\pm$ 0.002 & 0.667 $\pm$ 0.018 \\
UWaveGestureLibraryY & 896 / 3582 & 315 / 8 & 0.541 $\pm$ 0.018 & 0.661 $\pm$ 0.002 & 0.617 $\pm$ 0.023 \\
UWaveGestureLibraryZ & 896 / 3582 & 315 / 8 & 0.531 $\pm$ 0.027 & 0.657 $\pm$ 0.007 & 0.576 $\pm$ 0.011 \\
Wafer & 1000 / 6164 & 152 / 2 & 0.963 $\pm$ 0.006 & 0.982 $\pm$ 0.002 & 0.964 $\pm$ 0.010 \\
Wine & 57 / 54 & 234 / 2 & 0.650 $\pm$ 0.024 & 0.783 $\pm$ 0.020 & 0.872 $\pm$ 0.033 \\
WordSynonyms & 267 / 638 & 270 / 25 & 0.188 $\pm$ 0.022 & 0.378 $\pm$ 0.012 & 0.275 $\pm$ 0.025 \\
Worms & 181 / 77 & 900 / 5 & 0.370 $\pm$ 0.033 & 0.450 $\pm$ 0.020 & 0.419 $\pm$ 0.022 \\
WormsTwoClass & 181 / 77 & 900 / 2 & 0.537 $\pm$ 0.021 & 0.568 $\pm$ 0.020 & 0.575 $\pm$ 0.031 \\
Yoga & 300 / 3000 & 426 / 2 & 0.731 $\pm$ 0.054 & 0.840 $\pm$ 0.006 & 0.721 $\pm$ 0.022 \\
ACSF1 & 100 / 100 & 1460 / 10 & 0.525 $\pm$ 0.057 & 0.674 $\pm$ 0.031 & 0.636 $\pm$ 0.051 \\
BME & 30 / 150 & 128 / 3 & 0.368 $\pm$ 0.040 & 0.953 $\pm$ 0.006 & 0.957 $\pm$ 0.007 \\
Chinatown & 20 / 343 & 24 / 2 & 0.653 $\pm$ 0.129 & 0.960 $\pm$ 0.013 & 0.951 $\pm$ 0.007 \\
Crop & 7200 / 16800 & 46 / 24 & 0.453 $\pm$ 0.022 & 0.714 $\pm$ 0.005 & 0.442 $\pm$ 0.022 \\
DodgerLoopDay & 78 / 80 & 288 / 7 & 0.155 $\pm$ 0.030 & 0.549 $\pm$ 0.026 & 0.544 $\pm$ 0.027 \\
DodgerLoopGame & 20 / 138 & 288 / 2 & 0.479 $\pm$ 0.016 & 0.648 $\pm$ 0.020 & 0.749 $\pm$ 0.004 \\
DodgerLoopWeekend & 20 / 138 & 288 / 2 & 0.470 $\pm$ 0.098 & 0.708 $\pm$ 0.049 & 0.565 $\pm$ 0.223 \\
EOGHorizontalSignal & 362 / 362 & 1250 / 12 & 0.381 $\pm$ 0.051 & 0.427 $\pm$ 0.015 & 0.431 $\pm$ 0.017 \\
EOGVerticalSignal & 362 / 362 & 1250 / 12 & 0.210 $\pm$ 0.023 & 0.323 $\pm$ 0.021 & 0.351 $\pm$ 0.039 \\
EthanolLevel & 504 / 500 & 1751 / 4 & 0.355 $\pm$ 0.030 & 0.586 $\pm$ 0.007 & 0.648 $\pm$ 0.039 \\
FreezerRegularTrain & 150 / 2850 & 301 / 2 & 0.946 $\pm$ 0.027 & 0.989 $\pm$ 0.001 & 0.967 $\pm$ 0.006 \\
FreezerSmallTrain & 28 / 2850 & 301 / 2 & 0.728 $\pm$ 0.023 & 0.819 $\pm$ 0.006 & 0.804 $\pm$ 0.004 \\
Fungi & 18 / 186 & 201 / 18 & 0.080 $\pm$ 0.030 & 0.536 $\pm$ 0.014 & 0.006 $\pm$ 0.003 \\
GunPointAgeSpan & 135 / 316 & 150 / 2 & 0.932 $\pm$ 0.028 & 0.964 $\pm$ 0.001 & 0.851 $\pm$ 0.055 \\
GunPointMaleVersusFemale & 135 / 316 & 150 / 2 & 0.984 $\pm$ 0.009 & 1.000 $\pm$ 0.000 & 0.932 $\pm$ 0.022 \\
GunPointOldVersusYoung & 136 / 315 & 150 / 2 & 0.999 $\pm$ 0.001 & 1.000 $\pm$ 0.000 & 1.000 $\pm$ 0.000 \\
HouseTwenty & 40 / 119 & 2000 / 2 & 0.727 $\pm$ 0.020 & 0.748 $\pm$ 0.007 & 0.738 $\pm$ 0.014 \\
InsectEPGRegularTrain & 62 / 249 & 601 / 3 & 1.000 $\pm$ 0.000 & 1.000 $\pm$ 0.000 & 1.000 $\pm$ 0.000 \\
InsectEPGSmallTrain & 17 / 249 & 601 / 3 & 0.732 $\pm$ 0.155 & 1.000 $\pm$ 0.000 & 1.000 $\pm$ 0.000 \\
MelbournePedestrian & 1194 / 2439 & 24 / 10 & 0.764 $\pm$ 0.032 & 0.895 $\pm$ 0.007 & 0.642 $\pm$ 0.046 \\
MixedShapesRegularTrain & 500 / 2425 & 1024 / 5 & 0.825 $\pm$ 0.024 & 0.899 $\pm$ 0.002 & 0.872 $\pm$ 0.014 \\
MixedShapesSmallTrain & 100 / 2425 & 1024 / 5 & 0.752 $\pm$ 0.028 & 0.808 $\pm$ 0.003 & 0.803 $\pm$ 0.014 \\
PigAirwayPressure & 104 / 208 & 2000 / 52 & 0.037 $\pm$ 0.010 & 0.080 $\pm$ 0.012 & 0.091 $\pm$ 0.010 \\
PigArtPressure & 104 / 208 & 2000 / 52 & 0.101 $\pm$ 0.014 & 0.157 $\pm$ 0.012 & 0.165 $\pm$ 0.021 \\
PigCVP & 104 / 208 & 2000 / 52 & 0.058 $\pm$ 0.021 & 0.085 $\pm$ 0.009 & 0.062 $\pm$ 0.006 \\
PowerCons & 180 / 180 & 144 / 2 & 0.918 $\pm$ 0.040 & 0.972 $\pm$ 0.005 & 0.999 $\pm$ 0.002 \\
Rock & 20 / 50 & 2844 / 4 & 0.266 $\pm$ 0.058 & 0.605 $\pm$ 0.028 & 0.581 $\pm$ 0.017 \\
SemgHandGenderCh2 & 300 / 600 & 1500 / 2 & 0.732 $\pm$ 0.025 & 0.873 $\pm$ 0.011 & 0.790 $\pm$ 0.016 \\
SemgHandMovementCh2 & 450 / 450 & 1500 / 6 & 0.411 $\pm$ 0.021 & 0.590 $\pm$ 0.010 & 0.504 $\pm$ 0.016 \\
SemgHandSubjectCh2 & 450 / 450 & 1500 / 5 & 0.484 $\pm$ 0.056 & 0.873 $\pm$ 0.004 & 0.714 $\pm$ 0.056 \\
SmoothSubspace & 150 / 150 & 15 / 3 & 0.862 $\pm$ 0.048 & 0.974 $\pm$ 0.003 & 0.859 $\pm$ 0.009 \\
UMD & 36 / 144 & 150 / 3 & 0.369 $\pm$ 0.071 & 0.795 $\pm$ 0.020 & 0.911 $\pm$ 0.011 \\
\end{longtable}}

\end{document}